\documentclass[letterpaper, 10 pt, journal, twoside]{IEEEtran}

\IEEEoverridecommandlockouts                              % This command is only needed if 
\usepackage{multicol}
\usepackage[bookmarks=true]{hyperref}

\usepackage{float}
\usepackage{comment}
\usepackage{array}
\usepackage{graphicx}
\usepackage{booktabs}
\usepackage{multirow}

\usepackage{amsmath}								
\usepackage{amssymb}
\usepackage{latexsym}
\usepackage{amsthm}
\usepackage{bm}
\usepackage{commath}
\usepackage{float}
\usepackage{units}

\usepackage{tikz}
\usetikzlibrary{arrows,shapes,trees,calc}
\usetikzlibrary{backgrounds}
\usepackage{pgfplots}
\pgfplotsset{compat=1.14}
\usepgfplotslibrary{polar}
\usepgfplotslibrary{patchplots}

\usepackage{tikz-3dplot} %requires 3dplot.sty to be in same directory, or in your LaTeX installation
\newtheorem{definition}{Definition}[section]

\definecolor{lightyellow}{RGB}{255,236,132}
\definecolor{lightgreen}{RGB}{161,239,10}
\definecolor{darkgreen}{RGB}{61,124,68}
\definecolor{lightblue}{RGB}{72,131,219}
\definecolor{darkblue}{RGB}{39,63,186}
\definecolor{plgreen}{RGB}{27,158,119}
\definecolor{plorange}{RGB}{217,95,2}
\definecolor{plpurple}{RGB}{117,112,179}
\definecolor{plpink}{RGB}{231,41,138}

\usepackage{color}  % For Highlighting

\newcommand{\David}[1]{\textcolor{red}{(#1)}}

\newcommand{\Dan}[1]{\textcolor{magenta}{(#1)}}

\newcommand{\bmx}{\begin{bmatrix}}
\newcommand{\emx}{\end{bmatrix}}

\usepackage{marginnote}
\newcommand{\revcomment}[2]{\textcolor{red}{#2} \marginnote{\##1}} % Include reference commands
\renewcommand{\revcomment}[2]{#2}   % removes coloring of edited sections, i.e. makes doc look normal (comment this line out for highlighted edits)

\title{\LARGE \bf
Force Generation by Parallel Combinations of \\ Fiber-Reinforced Fluid-Driven Actuators
}

\author{Daniel Bruder$^{1}$, %<-this % stops a space
        Audrey Sedal$^{1}$, %
        Ram Vasudevan$^{1}$, %
        and C. David Remy$^{1}$%
\thanks{Manuscript received: February, 25, 2018; Revised May, 29, 2018; Accepted July, 9, 2018.}    % use only for final RAL version
\thanks{This paper was recommended for publication by Editor Yu Sun upon evaluation of the Associate Editor and Reviewers' comments.
*This material is supported by the Toyota Research Institute, and is based upon work supported by the National Science Foundation Graduate Research Fellowship Program under Grant No. 1256260 DGE. Any opinions, findings, and conclusions or recommendations expressed in this material are those of the author(s) and do not necessarily reflect the views of the National Science Foundation.}% <-this % stops a space
\thanks{$^{1}$ The authors are with the Mechanical Engineering Department at the 
        University of Michigan, Ann Arbor, MI 48109, USA
        {\tt\small \{bruderd, asedal, ramv, cdremy\}@umich.edu}}%
\thanks{Digital Object Identifier (DOI): see top of this page.}
}

\begin{document}

\maketitle
% \thispagestyle{empty}         % commented for final RAL version
% \pagestyle{empty}             % commented for final RAL version

% Paper headers (for final RAL submission)
\markboth{IEEE Robotics and Automation Letters. Preprint Version. Accepted July, 2018}
{Bruder \MakeLowercase{\textit{et al.}}: Force Generation by Parallel Combinations of Fluid-Driven Actuators}  % Use only for final RAL version

\begin{abstract}
% context
The compliant structure of soft robotic systems enables a variety of novel capabilities in comparison to traditional rigid-bodied robots.
A subclass of soft fluid-driven actuators known as fiber-reinforced elastomeric enclosures (FREEs) is particularly well suited as actuators for these types of systems.
FREEs are inherently soft and can impart spatial forces without imposing a rigid structure.
Furthermore, they can be configured to produce a large variety of force and moment combinations.
In this paper we explore the potential of combining multiple differently configured FREEs in parallel to achieve fully controllable multi-dimensional soft actuation.
% object
To this end, we propose a novel methodology to represent and calculate the generalized forces generated by soft actuators as a function of their internal pressure.
This methodology relies on the notion of a state dependent \emph{fluid Jacobian} that yields a linear expression for force. We employ this concept to construct the set of all possible forces that can be generated by a soft system in a given state.
This \emph{force zonotope} can be used to inform the design and control of parallel combinations of soft actuators.  
% findings
The approach is verified experimentally with the parallel combination of three carefully designed actuators constrained to a 2DOF test platform.
The force predictions matched measured values with a root-mean-square error of less than \unit[1.5]{N} force and \unit[$\mathbf{8 \times 10^{-3}}$]{Nm} moment, demonstrating the utility of the presented methodology.  
% conclusion

% perspectives

% composed of an elastomer tube constrained by inextensible fibers

% Soft fluid-driven actuators are well suited to this task because they are capable of imparting generalized spacial forces without imposing rigid structure. A subclass of these actuators, known as Fiber Reinforced Elastomeric Enclosures (FREEs), are particularly suitable for their configurability... \Dan{sentence is work in progress}.

% a physical realization of
\end{abstract}

% Keywords appear just beneath the abstract. Use only for final RAL version.

\IEEEpeerreviewmaketitle

% \input{outline}

% Input sections here
\section{Introduction}
\label{sec:introduction}

%% What are soft robots, why are they useful?
\IEEEPARstart{S}{oft} robotic systems have the potential to offer capabilities that go far beyond the repertoire of traditional rigid-bodied robots.
Their compliant structure allows them to adapt their overall shape to navigate unstructured environments, to safely work alongside humans, to manipulate delicate goods, and to absorb impacts without damage \cite{majidi2014soft}. 
Furthermore, soft robotic technology enables designs that are cheap to manufacture and that can be completely encapsulated to provide robust protection from the environment.

\begin{figure}
\centering

\def\picScale{0.08}    % define variable for scaling all pictures evenly
\def\colWidth{0.5\linewidth}

\begin{tikzpicture}
\matrix [row sep=0.25cm, column sep=0cm, style={align=center}] (my matrix) at (0,0) %(2,1)
{
\node[style={anchor=center}] (FREEhand) {\includegraphics[width=0.85\linewidth]{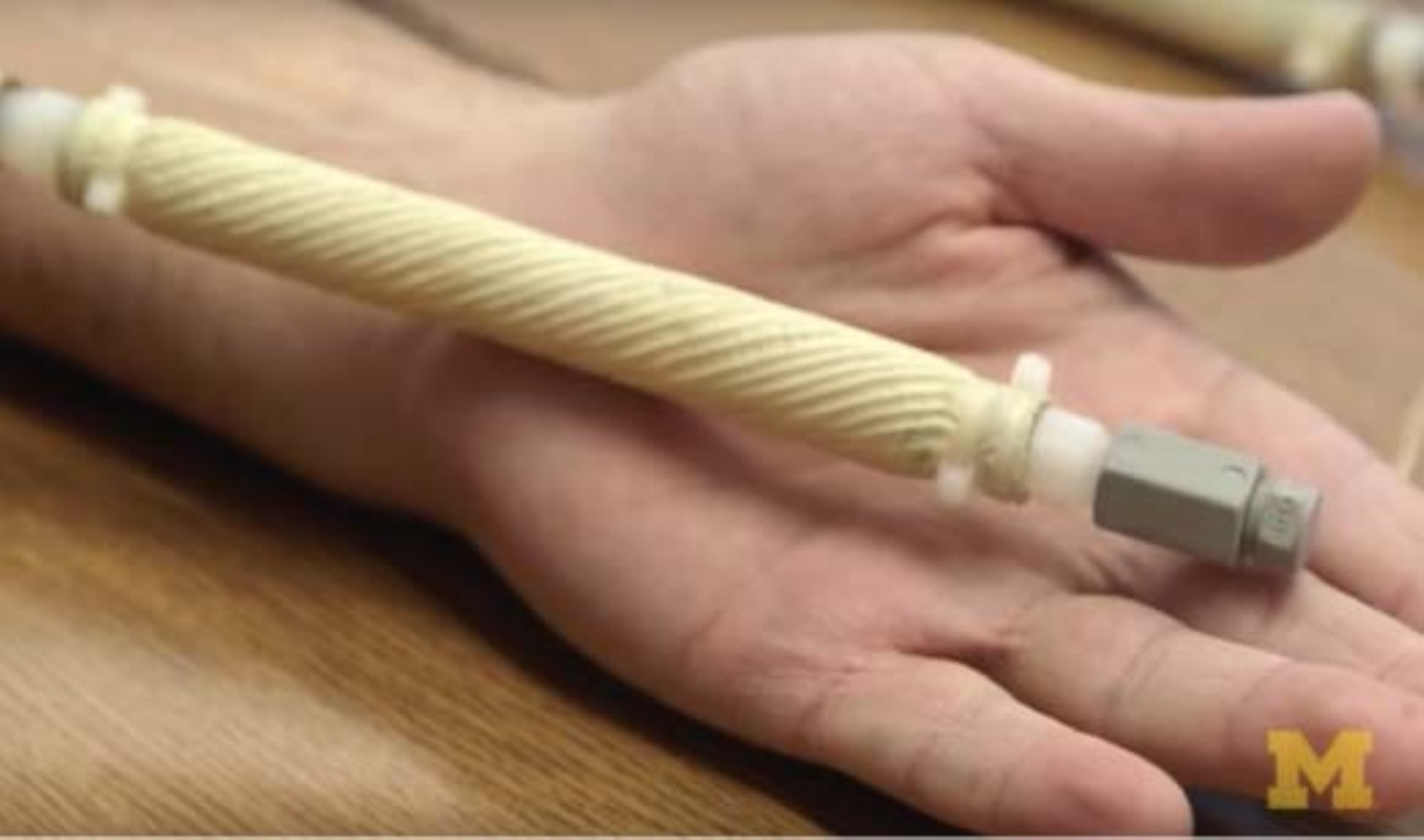}}; %\fill[blue] (0,0) circle (2pt);
\\
\node[style={anchor=center}] (rigid_v_soft) {\includegraphics[width=0.75\linewidth]{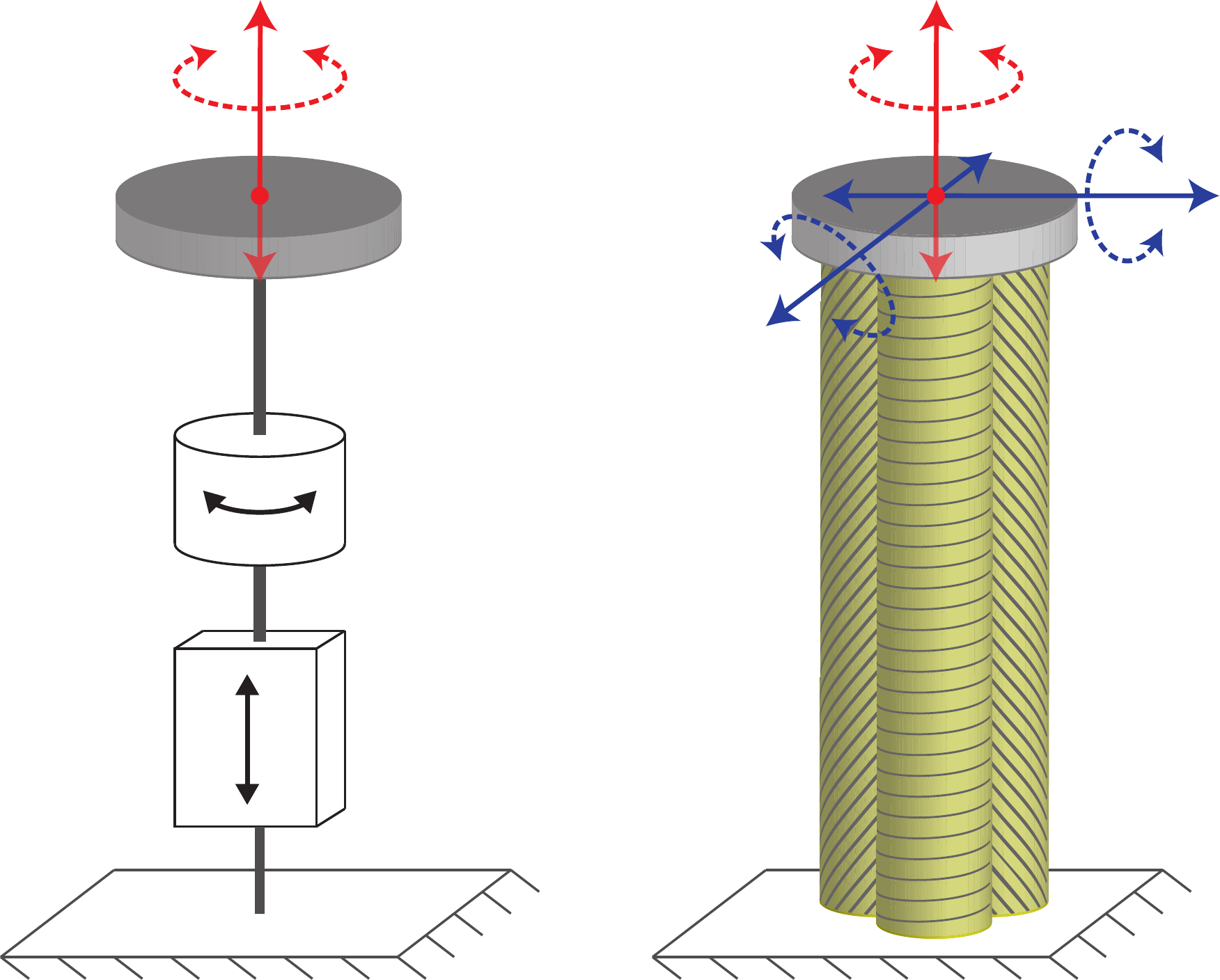}}; %\fill[blue] (0,0) circle (2pt);
\\
};
\node[above] (FREEhand) at ($ (FREEhand.south west)  !0.05! (FREEhand.south east) + (0, 0.1)$) {(a)};
\node[below] (a) at ($ (rigid_v_soft.south west) !0.20! (rigid_v_soft.south east) $) {(b)};
\node[below] (b) at ($ (rigid_v_soft.south west) !0.75! (rigid_v_soft.south east) $) {(c)};
\end{tikzpicture}

% \begin{tikzpicture} %[every node/.style={draw=black}]
% % \draw[help lines] (0,0) grid (4,2);
% \matrix [row sep=0cm, column sep=0cm, style={align=center}] (my matrix) at (0,0) %(2,1)
% {
% \node[style={anchor=center}] {\includegraphics[width=\colWidth]{figures/photos/labFREEs3.jpg}}; %\fill[blue] (0,0) circle (2pt)
% &
% \node[style={anchor=center}] {\includegraphics[width=\colWidth, height=160pt]{figures/stewartRender.png}}; %\fill[blue] (0,0) circle (2pt);
% \\
% };

% %\node[style={anchor=center}] at (0,-5) (FREEstate) {\includegraphics[width=0.7\linewidth]{figures/FREEstate_noLabels2.pdf}};

% \end{tikzpicture}

\caption{\revcomment{2.3}{(a) A fiber-reinforced elastomerc enclosure (FREE) is a soft fluid-driven actuator composed of an elastomer tube with fibers wound around it to impose specific deformations under an increase in volume, such as extension and torsion. (b) A linear actuator and motor combined in \emph{series} has the ability to generate 2 dimensional forces at the end effector (shown in red), but is constrained to motions only in the directions of these forces. (b) Three FREEs combined in \emph{parallel} can generate the same 2 dimensional forces at the end effector (shown in red), without imposing kinematic constraints that prohibit motion in other directions (shown in blue).}}

% \caption{A fiber-reinforced elastomeric enclosure (FREE) (top) is a soft fluid-driven actuator composed of an elastomer tube with fibers wound around it to impose deformation in specific directions upon pressurization, such as extension and torsion. \revcomment{2.3}{In this paper we explore the potential of combining multiple FREEs in parallel to generate fully controllable multi-dimensional spacial forces}, such as in a parallel arrangement around a flexible spine element (bottom-left), or a Stewart Platform arrangement (bottom-right).}

\label{fig:overview}
\end{figure}
   % introductory figure

%% Fluid-driven actuators are a good choice for soft robots
To obtain all of these advantages, it is crucial that every component of a soft robotic system is made in a compliant fashion.
As a result, many soft robotic systems are actuated by fluid-driven soft actuators that can produce forces without imposing a rigid structure \cite{grissom2006design, hawkes2017soft, marchese2014autonomous, tolley2014resilient}. 
In these actuators, a pressurized fluid such as water or air creates a targeted deformation of a soft structure that encloses a fluid-filled cavity. 
To achieve a specific type and direction of deformation, and not merely a homogeneous expansion, the stiffness of the soft structure is pattered in a specific way by adding reinforcing elements such as fibers, beams, or plates \cite{galloway2013mechanically, marchese2015recipe, rus2015design}. 
Examples of these actuators include bellows \cite{pridham1967bellows} McKibben actuators \cite{tondu2012modelling}, and pneu-nets \cite{mosadegh2014pneumatic}.
%, either by the use of inhomogeneous \David{or: `anisotropic'?} materials (cite whitesides, Rus, others from Rus review) or

%% FREEs are great because they are fluid-driven and customizable
A particularly promising type of soft fluid-driven actuator is the fiber-reinforced elasomeric enclosure (FREE) \cite{bishop2015design, krishnan2012evaluating, bishop2013force}, \revcomment{1.7}{also known as the fiber-reinforced soft actuator (FRSA) \cite{galloway2013mechanically, connolly2015mechanical, connolly2017automatic}}.
A FREE consists of a fluid-filled elastomeric tube wound with reinforcing fibers that pattern its stiffness to yield a desired mode and direction of deformation upon pressurization. %(Fig.~\ref{fig:FREEhand}). 
By changing the angles and arrangement of these fibers, a FREE can be customized to yield a large variety of desired deformations and forces \cite{bishop2015design}. 
\revcomment{1.2, 3.5}{This customizability combined with their flexibility and tube-like shape makes these actuators well suited for applications such as a pipe inspection \cite{connolly2015mechanical}, catheter devices \cite{gilbertson2016soft}, or continuum manipulators \cite{grissom2006design}.}

%% Basic argument: Rigid robots: structure constrained, actuators exert forces at joints to move structure,  Soft robots: structure unconstrained, actuators impose forces/constraints to move structure
Due to their soft and deformable structure, FREEs (and fluid driven soft actuators in general) differ in a fundamental way from traditional actuators.
An electric motor, for example, essentially combines a kinematic constraint (the rotation axis of the motor, which is physically defined by a pair of bearings) with a force generating element (the electromagnetic forces, which create the motor torque).
Since the motion of such an actuator is inherently limited to one dimension, multiple actuator stages are typically combined in \emph{series} to achieve multi degree of freedom (DOF) motions (Fig.~\ref{fig:overview}b). 
This is the prevalent design, for example, in industrial robotic arms.
In contrast, in a soft actuator, the force generating element is not supported by any physical kinematic constraints.
The actuator produces a spatial force without constraining the motion to happen exclusively in the direction of this force.
Because of this, soft actuators are particularly well suited to be combined in \emph{parallel}, where the forces of the individual actuators are superimposed to generate a multi-dimensional spatial force (Fig.~\ref{fig:overview}c).
Such parallel combinations enable particularly compact designs of multi-DOF motion stages.
% \Dan{EDIT THIS: They have equivalents in the world of traditional robotic systems, such as the well-known Stewart Platform.
% However, in such systems, the complexity is much higher, as each individual actuator needs to be combined with five additional joints to overcome the inherent kinematic constraints.}

%\begin{figure}
%    \centering
%    \includegraphics[width=\linewidth]{figures/FREEhand.png}
%    \caption{Possibly just a placeholder, not sure if we can use this image... \David{If you want to make Fig. 1 the overview figure, i would include 4 panels: 1) picture of free (current Fig 1), 2) drawing of free (current fig 3), 3) a parallel combination of frees (i.e., your system) and 4) maybe a steward platform to contrast.}}
%    \label{fig:FREEhand}
%\end{figure}

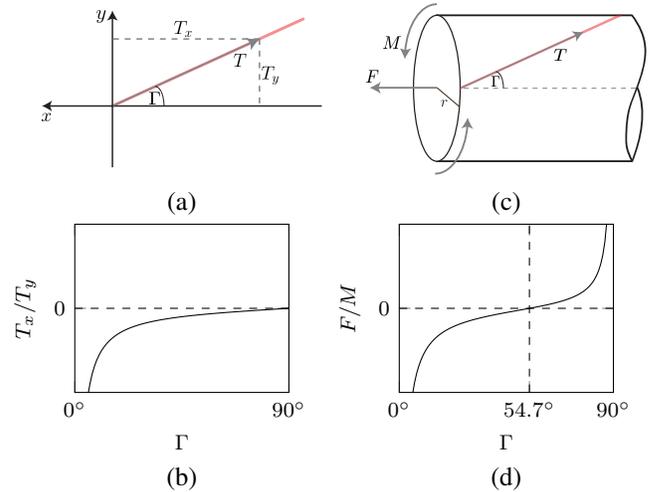
\begin{figure}
\centering

\begin{tikzpicture} %[every node/.style={draw=black}]
% \draw[help lines] (0,0) grid (4,2);
\matrix [row sep=0cm, column sep=0cm, style={align=center}] (my matrix) at (2,1)
{
\node[style={anchor=center}] {\includegraphics[width=0.42\linewidth]{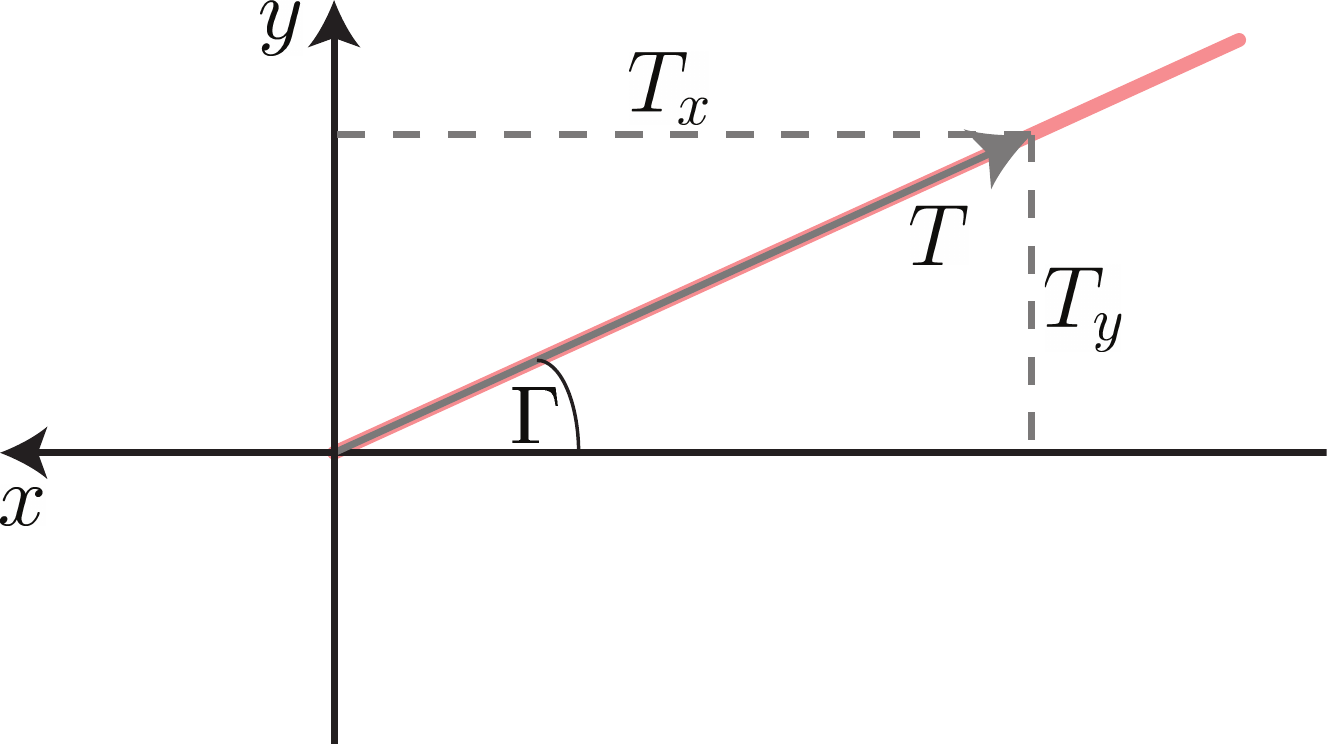}}; %\fill[blue] (0,0) circle (2pt);
&
\node[style={anchor=center}] {\includegraphics[width=0.42\linewidth]{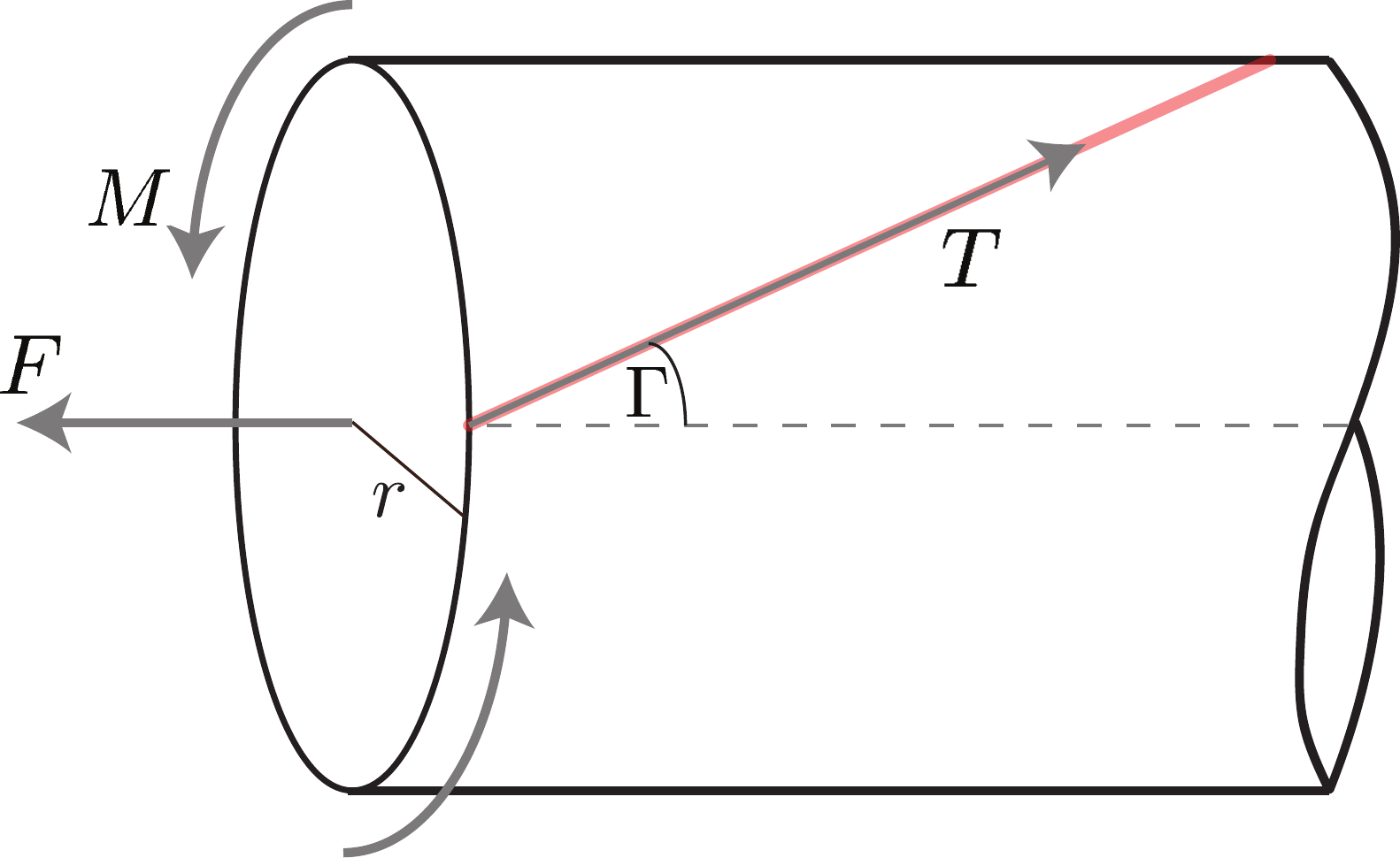}}; %\fill[blue] (0,0) circle (2pt);

\\
\node (a) {(a)}; & \node (c) {(c)};
\\

\begin{axis}[
    xlabel={\footnotesize{$\Gamma$}},
    ylabel={\footnotesize{$T_x/T_y$}},
    ymin=-10, ymax=10, ytick={0}, ylabel near ticks,
    xmin=0, xmax=90, xtick={0,90}, xticklabel=$\pgfmathprintnumber{\tick}^\circ$, xlabel near ticks,
    tick label style={font=\footnotesize},
    width=0.5\linewidth,
    anchor=center,
]
    \addplot [domain=0:90, dashed] {0};
    \addplot [domain=1:90, samples=100] {-cot(x)};
\end{axis};
& 
\begin{axis}[
    xlabel={\footnotesize{$\Gamma$}},
    ylabel={\footnotesize{$F/M$}},
    ymin=-10, ymax=10, ytick={0}, ylabel near ticks,
    xmin=0, xmax=90, xtick={0,54.7,90}, xticklabel=$\pgfmathprintnumber{\tick}^\circ$, xlabel near ticks,
    tick label style={font=\footnotesize},
    width=0.5\linewidth,
    anchor=center,
]
    \addplot [domain=0:90, dashed] {0};
    \addplot [domain=1:90, samples=100] {(1-2*cot(x)^2) / (2*cot(x))};
    \addplot [mark=none, dashed] coordinates {(54.7,-10) (54.7,10)};
\end{axis};
%\fill[blue] (0,0) circle (2pt);

\\
\node (b) {(b)}; & \node (d) {(d)};
\\
};
\end{tikzpicture}

\caption{By changing the fiber angle of a FREE, it can be configured to produce a large range of force/torque combinations. To understand this, we consider how (a) the angle at which a fiber is pulled in a plane affects (b) the ratio between the $x$ and $y$ components of the pulling force. Similarly, by (c) wrapping the plane into a cylinder and accounting for an internal pressure force pushing out on the endcap, the (d) ratio between axial force and twisting moment can be arbitrarily set by changing the fiber angle.}
\label{fig:FMratios}
\end{figure}

%% Adaptability of FREEs
One of the benefits of using FREEs as actuators is their mechanical programmability.
% , which makes them particularly well suited for parallel actuator combinations. 
By changing the fiber angle, $\Gamma$, of a single set of evenly distributed parallel fibers (Fig.~\ref{fig:overview}), a FREE can be configured to produce a variety of combinations of forces along its main axis as well as twisting moments about this axis.
The principle behind this adaptability can best be understood by considering a fiber being pulled in a plane (Fig.~\ref{fig:FMratios}a).
For a given fiber force, $T$, the ratio between the $x$ and $y$ components of $T$ is determined by the angle $\Gamma$ (Fig.~\ref{fig:FMratios}b). 
In the case of a FREE, this plane is wrapped into a cylinder (Fig.~\ref{fig:FMratios}c). 
Now the $x$-component of the force pulls along the direction of the central axis of the cylinder, while the $y$-component exerts the twisting moment.
Additional effects, such as the changing radius of the FREE and the fluid pressure acting on its end-caps, make the process a bit more complicated \cite{bruder2017model}, but the ratio between the resulting axial force and twisting moment is fully determined by the fiber angle $\Gamma$ (Fig.~\ref{fig:FMratios}d).

\revcomment{2.2, 2.9}{
A number of researchers have developed ways of modeling fiber-reinforced actuators. Krishnan et. al. characterized the range of achievable motions based on fiber angles \cite{krishnan2012evaluating}, exploring beyond the scope of previous literature which had focused on McKibben actuators only \cite{tondu2012modelling}. Bishop-Moser et. al. formalized a geometry-based kinematic model of FREEs \cite{bishop2015design}, which was subsequently codified and expanded upon by others \cite{felt2018closed, krishnan2015kinematics}. Connolly et. al. took another approach, using finite element methods to predict the motions of fiber-reinforced actuators \cite{connolly2015mechanical}. Bishop-Moser \cite{bishop2013force}, Bruder \cite{bruder2017model}, and Sedal \cite{sedal2017constitutive}, introduced force prediction models based upon the principle of virtual work, force balance, and continuum mechanics, respectively. Furthermore, several papers have been written describing kinematic models of parallel combinations of fiber-reinforced actuators \cite{bishop2012parallel, bishop2012parallelsynth, pritts2004design}.
}

This paper explores the potential of combining different types of FREE actuators in parallel to achieve fully controllable spatial forces, and provides the first generalized kinetic model of parallel combinations of FREEs (to the authors' knowledge).
This work thus expands on the existing literature regarding fiber-reinforced fluid-driven actuators, which has focused mainly on the kinematics or kinetics of individual actuators, or the kinematics of parallel combinations of actuators.
\revcomment{1.5}{While series combinations of FREEs may also be of interest for some applications, that is not the focus of this work.}
Here, we investigate how parallel combinations of FREEs can be configured to enable effective control of multi-DOF forces.
% Here, we study which parallel combinations and configurations of FREEs enable effective control of multi-DOF forces.
To this end, we present a novel way to represent and calculate actuator forces in terms of a state dependent fluid-Jacobian.
% This concept readily extends from a single soft actuator to parallel combinations of actuators.
Our design and modeling methodology is then employed and evaluated experimentally on a two degree of freedom test platform.

% This paper explores the potential of combining different types of FREE actuators in parallel to achieve fully controllable spatial forces.
% This work thus expands on the existing literature regarding fiber-reinforced fluid-driven actuators, which has focused mainly on the kinematics  \cite{bishop2015design, connolly2015mechanical, felt2018closed, krishnan2015kinematics} or kinetics \cite{bishop2013force, bruder2017model, sedal2017constitutive} of individual actuators, \revcomment{1.5}{applications of series combinations of actuators \cite{gilbertson2016soft}}, or the kinematics of parallel combinations of actuators \cite{bishop2012parallel, bishop2012parallelsynth}, \revcomment{2.2}{\cite{pritts2004design}}.
% Here, we study which combinations and configurations of FREEs enable effective control of multi-DOF forces.
% To this end, we present a novel way to represent and calculate actuator forces in terms of a state dependent fluid-Jacobian.
% This concept readily extends from a single soft actuator to parallel combinations of actuators.
% Our design and modeling methodology is employed and evaluated experimentally on a two degree of freedom test bench.

\section{Modeling}
\label{sec:singleActuator}
Our modeling approach for individual FREEs is based on the notion of a \emph{fluid Jacobian} $\bar{J}$, which maps the geometrical deformation of a soft actuator, or of a system of actuators, to a change in their volume. 
Under certain assumptions, the transpose of this Jacobian linearly maps the internal fluid pressure to the spatial forces that the actuator generates. 
One can  think of this fluid Jacobian as the soft and multi-dimensional equivalent of the cross section $A$ of a traditional pneumatic or hydraulic cylinder,
since this cross section similarly relates cylinder pressure to force, and piston movement to fluid displacement.

\subsection{Force Generation in a Single FREE}
The approach presented in this paper is enabled by a number of simplifying assumptions.
They are consistent with those made in prior work on the modeling and control of individual FREEs \cite{bishop2015design,bruder2017model}.
In particular, we assume that the fibers are inextensible and that they are uniformly  distributed  around  an elastomeric cavity  with negligible  wall thickness.
%The internal pressure in this cavity is uniform along its entire length.
Under these assumptions, a FREE can be modeled as a composition of an energy transforming element (the fibers) and an energy storing element (the compliance of the elastomer body). 
The generalized forces generated by each of these separate elements can be superimposed to characterize the net force $\vec{\tau}_{total}$ produced by the FREE:
\begin{align}
   \vec{\tau}_{\text{total}} &=  \vec{\tau} + \vec{\tau}_{\text{elast}}    \label{eq:netF}
\end{align}
where $\vec{\tau}$ and $\vec{\tau}_{\text{elast}}$ are the generalized forces and torques attributed to the fiber and elastomer, respectively.
In this work, we focus exclusively on the active general forces $\vec{\tau}$ that are generated by the fibers and that can be controlled by varying the pressure of the fluid.

The fluid Jacobian for such an actuator can be derived from the idea that the reinforcing fibers in a FREE create a kinematic constraint on the internal volume $V$ of the fluid cavity.
Without the reinforcing fibers, this cavity could expand freely, since it is made from soft material.
With the fibers, however, the volume is limited.
This limitation on volume depends on the mechanical parameters of the FREE (e.g., the relaxed fiber angle or fiber length) and on the current state of geometric deformation of the FREE.
This geometric state can be represented by a generalized vector of spatial deformations $\vec{q}$:  $V = V\left(\vec{q}\right)$.

The derivative of this volume yields an expression $\dot{V}$ for the volumetric flow into the actuator:
\begin{align}
    \dot{V} (\vec{q}, \dot{\vec{q}}) &= \frac{\partial V}{\partial t} = \frac{\partial V}{\partial \vec{q}} \frac{\partial \vec{q}}{\partial t } = \bar{J}_q (\vec{q}) \dot{\vec{q}},  \label{eq:Vdot_wJ}
\end{align}
where the fluid Jacobian is defined as $\bar{J}_{q}= \frac{\partial V}{\partial \vec{q}}$ with respect to the deformation $\vec{q}$.

It is important to note that $\vec{\tau}$ and $\dot{\vec{q}}$ live in dual spaces. 
That is, forces in $\vec{\tau}$ correspond to linear motion in $\dot{\vec{q}}$, and moments in $\vec{\tau}$ correspond to angular motion in $\dot{\vec{q}}$.
In the most general case, these are 6-dimensional vectors with three translations and three rotations.
In this case, $\bar{J}_q$ is a $1 \times 6$ matrix.

Energy conservation dictates that the mechanical power generated by the fibers must equal the fluid power that goes into the FREE:
\begin{align}
    P_{\text{mech}} = P_{\text{fluid}} = \dot{\vec{q}}^{\,T} \vec{\tau} &= \dot{V} p, 
    \label{eq:Pequiv}
\end{align}
where $p>0$ is the pressure of the fluid inside the FREE.
Substituting \eqref{eq:Vdot_wJ} into \eqref{eq:Pequiv}, we arrive at a linear expression that yields the forces produced by the FREE fibers as a function of deformation state and fluid pressure: 
\begin{align}
    \vec{\tau} (\vec{q}, p) &= \bar{J}_q^T (\vec{q}) p       \label{eq:fiberF}
\end{align}
The fluid Jacobian thus describes the generalized direction in which a FREE can produce forces and torques.
Since the input pressure $p$ needs to be strictly positive to avoid collapsing of the fluid cavity, forces can only be produced in the positive direction defined by $\bar{J}_q$.

\subsection{The Fluid Jacobian of a Cylindrical FREE}
The theoretical modeling approach outlined above can be applied independently of the exact geometry of a FREE and can even be extended to other classes of soft actuators.
However, to make the computation of the fluid Jacobian tractable, we rely on another common assumption for FREES: that they maintain the geometry of an ideal cylinder \cite{bishop2015design}.
This neglects the tapering of the actuator towards the end-caps and any potential bending along its main axis.

\begin{figure}
    \centering
    \begin{tikzpicture}
        \node (FREEstate) at (0,0)
            {\includegraphics[width=0.75\linewidth]{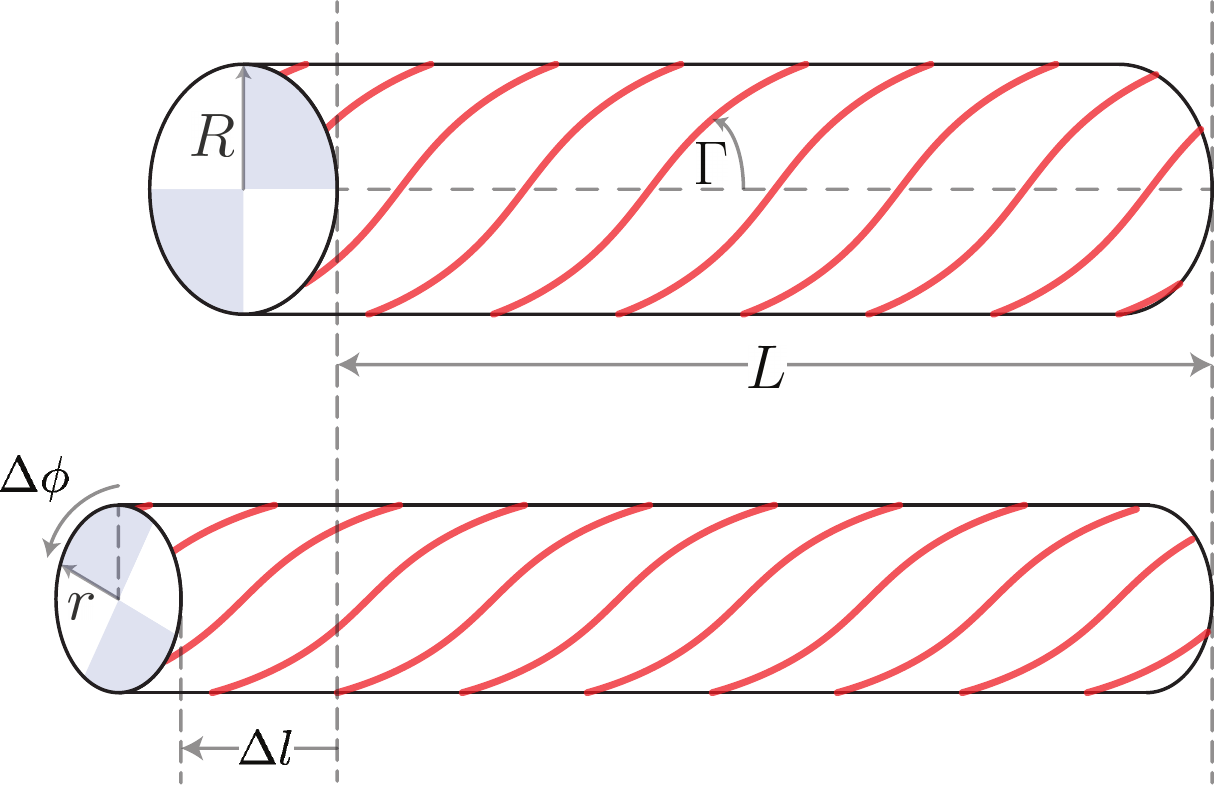}};
        \node[right] (a) at ($ (FREEstate.east) !0.5! (FREEstate.north east) $) {(a)};
        \node[right] (b) at ($ (FREEstate.east) !0.5! (FREEstate.south east) $) {(b)};
    \end{tikzpicture}
    \caption{Geometric parameters of an ideal cylindrical FREE in (a) the relaxed configuration where $\vec{q}=0$ (top), and (b) a deformed configuration where ${\vec{q}=\lbrack \Delta l \hspace{2pt} , \hspace{2pt} \Delta \phi \rbrack^T}$ (bottom).}
    \label{fig:FREEparams}
\end{figure}

An ideal cylindrical FREE in its relaxed configuration (i.e. when gauge fluid pressure is zero and no external loads are applied) can be described by a set of three parameters, $L$, $R$, and $\Gamma$, where $L$ represents the relaxed length of the FREE, $R$ represents the radius, and $\Gamma$ the fiber angle (Fig.~\ref{fig:FREEparams}). 
For notational convenience, we define two other quantities from these parameters:
\begin{align}
	B &= \left| \frac{L}{\cos{\Gamma}} \right| \\
	N &= - \frac{L}{2 \pi R} \tan{\Gamma}
\end{align}
where $B$ is the length of one of the FREE fibers, and $N$ is the total number of revolutions the fiber makes around the FREE in the relaxed configuration. 

\revcomment{2.4}{The assumption that a FREE is cylindrical with inextensible fiber-reinforcements implies that changes in its radius, length, and twist are coupled 
% via the right triangle relationship shown in Fig. \ref{fig:fiber}
. Therefore, its geometrical deformation can be fully defined in terms of just two parameters, a change in its length $\Delta l$ and a twist about its main axis $\Delta \phi$.}
These two values constitute the vector of generalized deformations $\vec{q} = \left[ \Delta l, \Delta \phi \right]^T$.
Consequently the generalized torque vector $\vec{\tau} = \left[ F, M \right]^T$ describes a force along the main axis, $F$, and a torsional moment about that axis, $M$.

From this, the length and radius of the deformed FREE can be computed according to:
\begin{align}
    l &= L + \Delta l \label{eq:l} \\
	r &= \frac{B}{\abs{2 \pi N + \Delta \phi}} \sqrt{1 - \left( \frac{L+\Delta l}{B} \right)^2}, \label{eq:r}
\end{align}
and we can express the volume as
\begin{align}
	V(\vec{q}) &= \pi r^2 l \notag \\ 
	&= \frac{\pi (L+\Delta l) (B^2 - (L+\Delta l)^2)}{(2 \pi N + \Delta \phi)^2}  \label{eq:V}.
\end{align}

With this, the fluid Jacobian evaluates to:
\begin{align}
    \bar{J}_q (\vec{q})
    &= \begin{bmatrix} 
		        \frac{\pi \left( B^2 - 3(L + \Delta l)^2 \right)}{(2 \pi N + \Delta \phi)^2} & \frac{2 \pi (L+\Delta l) \left( (L+\Delta l)^2 - B^2 \right)}{(2 \pi N + \Delta \phi)^3}
		\end{bmatrix}.    \label{eq:Jv}
\end{align}

% % THIS FIGURE WILL BE REMOVED FOR SPACE IF NEEDED, BUT I WANT TO KEEP IT IF POSSIBLE 
% \begin{figure}
%     \centering
%     \begin{tikzpicture}
%         \node (fiber) at (0,0)
%             {\includegraphics[width=0.85\linewidth]{figures/fiber_noLabels2.png}};
%         \node[below] (a) at ($ (fiber.south west) !0.08! (fiber.south east) $) {(a)};
%         \node[below] (b) at ($ (fiber.south west) !0.32! (fiber.south east) $) {(b)};
%         \node[below] (c) at ($ (fiber.south west) !0.75! (fiber.south east) $) {(c)};
%     \end{tikzpicture}
%     \caption{(a) A FREE contains many parallel fibers, all part of the same fiber family. (b) One isolated fiber forms a helical constraint. (c) The FREE radius $r$, twist $\Delta \phi$, and length change $\Delta l$ are geometrically related via the right triangle formed by unwinding a fiber and laying it flat.}
%     \label{fig:fiber}
% \end{figure}

\subsection{Parallel Combinations of FREEs}
\label{sec:parallelActuators}
\begin{figure}
    \centering
    \begin{tikzpicture}
        \node (diagram) at (0,0)
            {\includegraphics[width=1.0\linewidth]{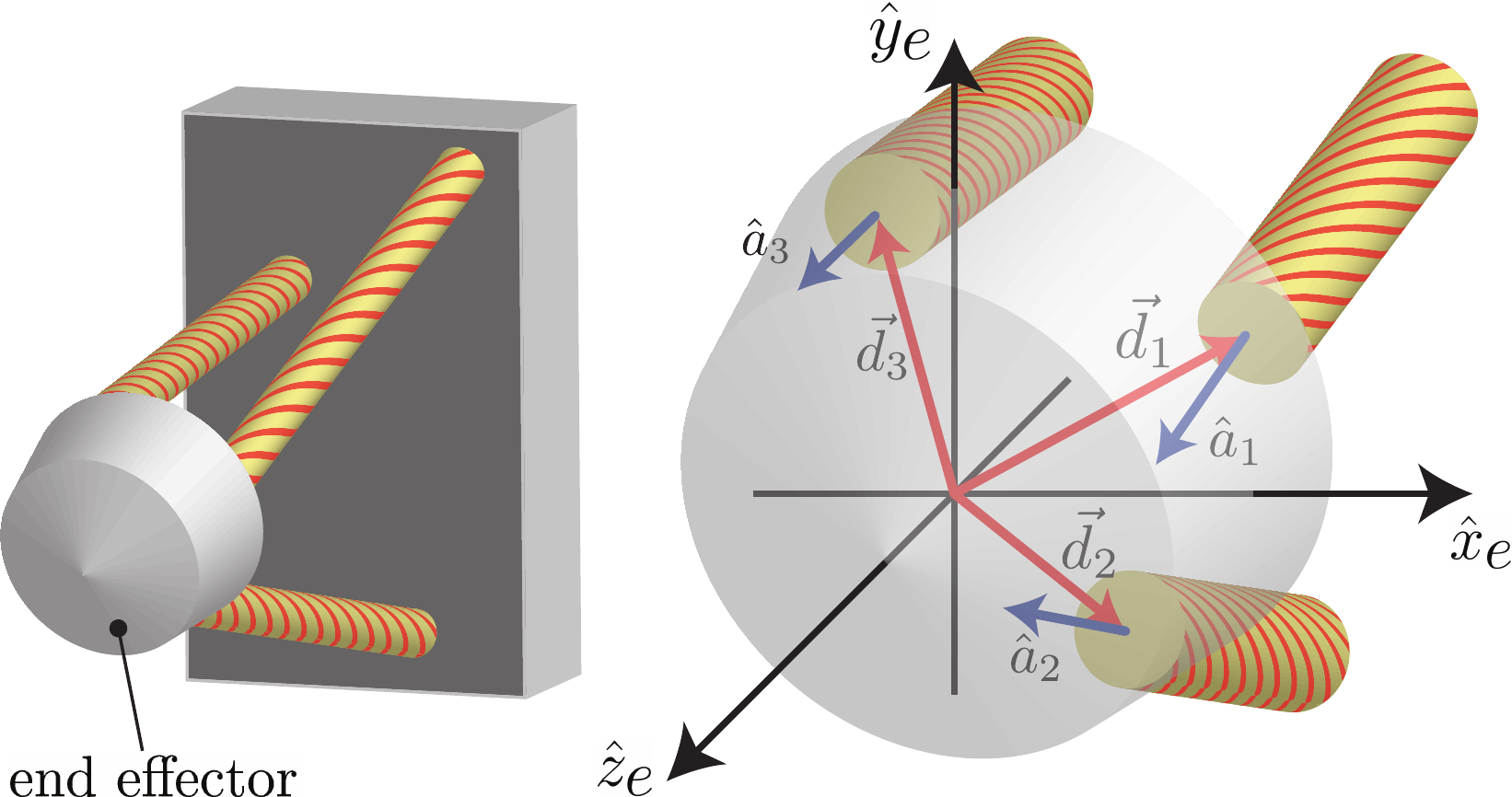}};
        \node[below] (a) at ($ (diagram.south west) !0.22! (diagram.south east) $) {(a)};
        \node[below] (b) at ($ (diagram.south west) !0.65! (diagram.south east) $) {(b)};
    \end{tikzpicture}
%    (a) An arbitrary combination of 3 FREEs connected to a common end effector. (b) A zoomed in view of the end effector with its local coordinate frame shown.}
    \caption{\revcomment{2.7}{(a) When multiple FREEs are attached in parallel to the same end effector, shown on the left, we must consider their relative positions and orientations. (b) This is determined by the vector $\vec{d}_i$ of the displacement of the attachment point of a FREE relative to the origin of the end effector frame, and by the unit vector $\hat{a}_i$ that is aligned with the FREE axis at the attachment point, shown in the zoomed-in view on the right.}}
    \label{fig:dp_defined}
\end{figure}

We can extend the concept of a fluid Jacobian to systems with multiple actuators that are mounted in parallel.
FREEs that are mounted in parallel each have one end attached to a common ground and the other rigidly attached to an end effector.
The position and orientation of this end effector is given by a deformation vector $\vec{x}$, and we assume that an inverse kinematics function allows the computation of the state $\vec{q}_i \left(\vec{x}\right)$ for each individual FREE.
To express forces and torques in a common reference frame, we define a body-fixed reference point and coordinate system that are attached to the end effector (Fig.~\ref{fig:dp_defined}). 
Expressed in these coordinates, a position vector ${\vec{d}_i = \begin{bmatrix} d_i^{\hat{x}_e} & d_i^{\hat{y}_e} & d_i^{\hat{z}_e} \end{bmatrix}^T}$ defines the point where the $i$th FREE is attached, and a unit vector ${\hat{a}_i = \begin{bmatrix} a_i^{\hat{x}_e} & a_i^{\hat{y}_e} & a_i^{\hat{z}_e} \end{bmatrix}^T}$, expresses the direction of the associated FREE axis.

Let $\vec{f}_i$ be the vector of general forces exerted by the $i$th FREE and expressed in end effector coordinates:
\begin{align}
    \vec{f}_i = \bmx F^{\hat{x}_e}_i & F^{\hat{y}_e}_i & F^{\hat{z}_e}_i & M^{\hat{x}_e}_i & M^{\hat{y}_e}_i & M^{\hat{z}_e}_i \emx^T,
\end{align}
where $F^{\hat{x}_e}_i$ is the component of force along the $x$-axis of the end effector frame and $M^{\hat{x}_e}_i$ is the moment about the $x$-axis of that frame.
The components of this vector can be computed from the axial force $F_i$ and twisting torque $M_i$ of the $i$th FREE as:
\begin{align}
    \bmx F^{\hat{x}_e}_i & F^{\hat{y}_e}_i & F^{\hat{z}_e}_i \emx^T &= \hat{a}_i F_i ,   \label{eq:Df}
\end{align}
and
\begin{align}
    \bmx M^{\hat{x}_e}_i & M^{\hat{y}_e}_i & M^{\hat{z}_e}_i \emx^T &= \lfloor \vec{d}_i \times \rfloor \hat{a}_i F_i + \hat{a}_i M_i,    \label{eq:Dm}
\end{align}
where $\lfloor \vec{d}_i \times \rfloor$ is the matrix notation for the cross-product with $\vec{d}_i$:
\begin{align}
    \lfloor \vec{d}_i \times \rfloor &= \begin{bmatrix} 0 & -d_i^{\hat{z}_e} & d_i^{\hat{y}_e} \\ d_i^{\hat{z}_e} & 0 & -d_i^{\hat{x}_e} \\ -d_i^{\hat{y}_e} & d_i^{\hat{x}_e} & 0 \end{bmatrix}. 
\end{align}
Combining \eqref{eq:Df} and \eqref{eq:Dm} into a single transformation yields:
\begin{align}
    \vec{f}_i &= \bar{\mathcal{D}}_i \vec{\tau}_i,  \label{eq:zetai}
\end{align}
where $\mathcal{D}_{i}$ is the $6 \times 2$ matrix:
\begin{align}
    \bar{\mathcal{D}}_i &= \begin{bmatrix}
                    \begin{bmatrix} \hat{a}_i & \vec{0}_{3\times1} \end{bmatrix} \vspace{5pt} \\ 
                    \begin{bmatrix} \lfloor \vec{d}_i \times \rfloor \hat{a}_i & \vec{0}_{3\times1} \end{bmatrix} + \begin{bmatrix} \vec{0}_{3\times1} & \hat{a}_i \end{bmatrix}
                    \end{bmatrix}.   \label{eq:D}
\end{align}

Since the actuators are mounted in parallel, the total force $\vec{f}$ is the sum of the individual forces of all $n$ FREEs connected to the end effector: 
\begin{align}
    \vec{f}(\vec{q}, \vec{p}) &= \sum_{i=1}^n \vec{f}_{i} = \sum_{i=1}^n \bar{\mathcal{D}}_i \bar{J}^T_{q, i} (\vec{q}_i) p_i = \sum_{i=1}^n \bar{J}^T_{x, i} (\vec{x}) p_i,
    \label{eq:zeta}
\end{align}
where $\bar{J}_{x, i} = \bar{J}_{q, i} \bar{\mathcal{D}}^T_i$ is the fluid Jacobian of an individual FREE expressed in end effector coordinates and the vector $\vec{p}$ contains the internal pressure values of all FREEs.

This can be written compactly in matrix notation as 
\begin{align}
    \vec{f} (\vec{x}, \vec{p}) &= \bar{J}^T_x (\vec{x}) \vec{p}, \label{DAVIDreallyLIKESthisEQUATION}
\end{align}
with the overall fluid Jacobian $\bar{J}_x$:
\begin{align}
    \bar{J}_x &= \bmx \bar{J}^T_{x, 1} & \bar{J}^T_{x, 2} & \cdots & \bar{J}^T_{x, n} \emx^T
\end{align}

\subsection{The Force Zonotope}
Equation \eqref{DAVIDreallyLIKESthisEQUATION} shows that the force capability of the parallel combination of multiple FREEs is a linear function of the pressures in the individual actuators.
Since the fluid Jacobian $\bar{J}_{x}$ depends on the end effector state $\vec{x}$, the ability of such a system to generate forces at the end effector will vary based on its displacement.
For some values of $\vec{x}$ it may be possible to generate spacial forces in multiple directions, while for others the span of possible forces may by narrower. 
The \emph{force zonotope} of a parallel combination of FREEs, similar to the force ellipsoid of rigid manipulators, 
% \cite{spong2008robot}
describes the set of forces that can be generated at the end effector with a bounded set of input pressures.

\begin{definition}[Force Zonotope]
    For a parallel combination of $n$ FREEs, the force zonotope is the set of active general 
%    torques and
    forces that can be generated in a specific end effector state, $\vec{x}$,
    \begin{align}
        \mathcal{Z}(\vec{x}) &= \left\{\bar{J}^T_x (\vec{x}) \vec{p} \, : \, p_i \in [0,p_i^\text{max}] \right\}     \label{eq:zonotope}
    \end{align}
    where $p_i^{\text{max}}$ is the maximum pressure allowed for the $i^{th}$ FREE.
\end{definition}

Note that the \emph{force zonotope} is the convex hull of all spacial forces generated when $p_i \in \{0, p_i^{\text{max}}\}$.
This makes it viable to compute using a convex hull algorithm. By calculating the zonotope over a range of states, it can be used to verify that a given parallel FREE design is capable of generating the desired forces over the range of its workspace. 
%In this way, the \emph{force zonotope} provides insight into how to utilize FREEs as actuators for robotic systems.

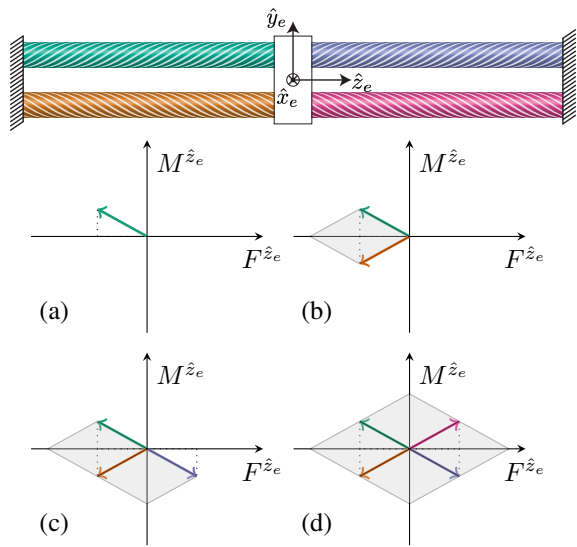
\begin{figure}
    \centering

\includegraphics[width=0.85\linewidth]{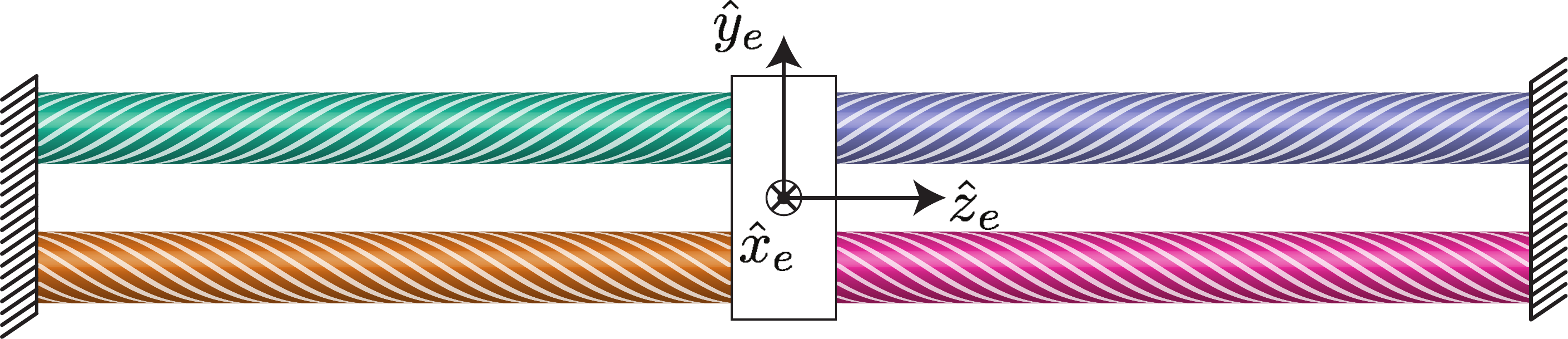}

\begin{tikzpicture}
\def\scl{0.45} % define scale variable for plots

% PRACTICE PLOT
\matrix [row sep=0.25cm, column sep=0cm, style={align=center}] (my matrix) at (0,0)
{

\begin{axis}[
    view={90}{0},
    axis lines=center,
    % axis equal image,
    xlabel={$M^{\hat{x}_e}$},
    ylabel={$F^{\hat{z}_e}$},
    zlabel={$M^{\hat{z}_e}$},
    ymin=-7, ymax=7, ytick={0}, %ylabel near ticks,
    xmin=-5, xmax=10, xtick={0}, %xticklabel=$\pgfmathprintnumber{\tick}^\circ$, xlabel near ticks, 
    zmin=-7, zmax=7, ztick={0}, %z dir=reverse,
    xlabel style={anchor=north}, ylabel style={anchor=north},
    scale=\scl,
    anchor=center,
    name=plot1,
    ]
    \def\pa{(-2,-3,2)}
    \def\pb{(2,-3,-2)}
    \def\pc{(2,3,-2)}
    \def\pd{(-2,3,2)}
    \addplot3[->, line width=1pt, plgreen] coordinates {(0,0,0) \pa};
%    \addplot3[->, line width=1pt, plorange] coordinates {(0,0,0) \pb};
%    \addplot3[->, line width=1pt, plpurple] coordinates {(0,0,0) \pc};
%    \addplot3[->, line width=1pt, plpink] coordinates {(0,0,0) \pd};
    % connector lines for perspective
    \addplot3[dotted] coordinates {(0,0,0) (-2,-3,0)};
    \addplot3[dotted] coordinates {(-2,-3,0) \pa}; 
%    \addplot3[dotted] coordinates {(0,0,0) (2,-3,0)};
%    \addplot3[dotted] coordinates {(2,-3,0) \pb};
%    \addplot3[dotted] coordinates {(0,0,0) (2,3,0)};
%    \addplot3[dotted] coordinates {(2,3,0) \pc};
%    \addplot3[dotted] coordinates {(0,0,0) (-2,3,0)};
%    \addplot3[dotted] coordinates {(-2,3,0) \pd};
    % faces of shape
%    \addplot3[patch, opacity=0.3, fill=black!20, faceted color=black, patch type=rectangle] 
%        coordinates {
%                    (0,0,0) \pa (0,-6,0) \pb
%                    (0,0,0) \pb (4,0,-4) \pc
%                    (0,0,0) \pc (0,6,0) \pd
%                    (0,0,0) \pd (-4,0,4) \pa
%                    };
\end{axis};

&

\begin{axis}[
    view={90}{0},
    axis lines=center,
    % axis equal image,
    xlabel={$M^{\hat{x}_e}$},
    ylabel={$F^{\hat{z}_e}$},
    zlabel={$M^{\hat{z}_e}$},
    ymin=-7, ymax=7, ytick={0}, %ylabel near ticks,
    xmin=-5, xmax=10, xtick={0}, %xticklabel=$\pgfmathprintnumber{\tick}^\circ$, xlabel near ticks, 
    zmin=-7, zmax=7, ztick={0}, %z dir=reverse,
    xlabel style={anchor=north}, ylabel style={anchor=north},
    scale=\scl,
    anchor=center,
    name=plot2,
    ]
    \def\pa{(-2,-3,2)}
    \def\pb{(2,-3,-2)}
    \def\pc{(2,3,-2)}
    \def\pd{(-2,3,2)}
    \addplot3[->, line width=1pt, plgreen] coordinates {(0,0,0) \pa};
    \addplot3[->, line width=1pt, plorange] coordinates {(0,0,0) \pb};
%    \addplot3[->, line width=1pt, plpurple] coordinates {(0,0,0) \pc};
%    \addplot3[->, line width=1pt, plpink] coordinates {(0,0,0) \pd};
    % connector lines for perspective
    \addplot3[dotted] coordinates {(0,0,0) (-2,-3,0)};
    \addplot3[dotted] coordinates {(-2,-3,0) \pa}; 
    \addplot3[dotted] coordinates {(0,0,0) (2,-3,0)};
    \addplot3[dotted] coordinates {(2,-3,0) \pb};
%    \addplot3[dotted] coordinates {(0,0,0) (2,3,0)};
%    \addplot3[dotted] coordinates {(2,3,0) \pc};
%    \addplot3[dotted] coordinates {(0,0,0) (-2,3,0)};
%    \addplot3[dotted] coordinates {(-2,3,0) \pd};
    % faces of shape
    \addplot3[patch, opacity=0.3, fill=black!20, faceted color=black, patch type=rectangle] 
        coordinates {
                    (0,0,0) \pa (0,-6,0) \pb
%                    (0,0,0) \pb (4,0,-4) \pc
%                    (0,0,0) \pc (0,6,0) \pd
%                    (0,0,0) \pd (-4,0,4) \pa
                    };
\end{axis};

\\

\begin{axis}[
    view={90}{0},
    axis lines=center,
    % axis equal image,
    xlabel={$M^{\hat{x}_e}$},
    ylabel={$F^{\hat{z}_e}$},
    zlabel={$M^{\hat{z}_e}$},
    ymin=-7, ymax=7, ytick={0}, %ylabel near ticks,
    xmin=-5, xmax=10, xtick={0}, %xticklabel=$\pgfmathprintnumber{\tick}^\circ$, xlabel near ticks, 
    zmin=-7, zmax=7, ztick={0}, %z dir=reverse,
    xlabel style={anchor=north}, ylabel style={anchor=north},
    scale=\scl,
    anchor=center,
    name=plot3,
    ]
    \def\pa{(-2,-3,2)}
    \def\pb{(2,-3,-2)}
    \def\pc{(2,3,-2)}
    \def\pd{(-2,3,2)}
    \addplot3[->, line width=1pt, plgreen] coordinates {(0,0,0) \pa};
    \addplot3[->, line width=1pt, plorange] coordinates {(0,0,0) \pb};
    \addplot3[->, line width=1pt, plpurple] coordinates {(0,0,0) \pc};
%    \addplot3[->, line width=1pt, plpink] coordinates {(0,0,0) \pd};
    % connector lines for perspective
    \addplot3[dotted] coordinates {(0,0,0) (-2,-3,0)};
    \addplot3[dotted] coordinates {(-2,-3,0) \pa}; 
    \addplot3[dotted] coordinates {(0,0,0) (2,-3,0)};
    \addplot3[dotted] coordinates {(2,-3,0) \pb};
    \addplot3[dotted] coordinates {(0,0,0) (2,3,0)};
    \addplot3[dotted] coordinates {(2,3,0) \pc};
%    \addplot3[dotted] coordinates {(0,0,0) (-2,3,0)};
%    \addplot3[dotted] coordinates {(-2,3,0) \pd};
    % faces of shape
    \addplot3[patch, opacity=0.3, fill=black!20, faceted color=black, patch type=rectangle] 
        coordinates {
                    (0,0,0) \pa (0,-6,0) \pb
                    (0,0,0) \pb (4,0,-4) \pc
%                    (0,0,0) \pc (0,6,0) \pd
%                    (0,0,0) \pd (-4,0,4) \pa
                    };
\end{axis};

&

\begin{axis}[
    view={90}{0},
    axis lines=center,
    % axis equal image,
    xlabel={$M^{\hat{x}_e}$},
    ylabel={$F^{\hat{z}_e}$},
    zlabel={$M^{\hat{z}_e}$},
    ymin=-7, ymax=7, ytick={0}, %ylabel near ticks,
    xmin=-5, xmax=10, xtick={0}, %xticklabel=$\pgfmathprintnumber{\tick}^\circ$, xlabel near ticks, 
    zmin=-7, zmax=7, ztick={0}, %z dir=reverse,
    xlabel style={anchor=north}, ylabel style={anchor=north},
    scale=\scl,
    anchor=center,
    name=plot4,
    ]
    \def\pa{(-2,-3,2)}
    \def\pb{(2,-3,-2)}
    \def\pc{(2,3,-2)}
    \def\pd{(-2,3,2)}
    \addplot3[->, line width=1pt, plgreen] coordinates {(0,0,0) \pa};
    \addplot3[->, line width=1pt, plorange] coordinates {(0,0,0) \pb};
    \addplot3[->, line width=1pt, plpurple] coordinates {(0,0,0) \pc};
    \addplot3[->, line width=1pt, plpink] coordinates {(0,0,0) \pd};
    % connector lines for perspective
    \addplot3[dotted] coordinates {(0,0,0) (-2,-3,0)};
    \addplot3[dotted] coordinates {(-2,-3,0) \pa}; 
    \addplot3[dotted] coordinates {(0,0,0) (2,-3,0)};
    \addplot3[dotted] coordinates {(2,-3,0) \pb};
    \addplot3[dotted] coordinates {(0,0,0) (2,3,0)};
    \addplot3[dotted] coordinates {(2,3,0) \pc};
    \addplot3[dotted] coordinates {(0,0,0) (-2,3,0)};
    \addplot3[dotted] coordinates {(-2,3,0) \pd};
    % faces of shape
    \addplot3[patch, opacity=0.3, fill=black!20, faceted color=black, patch type=rectangle] 
        coordinates {
                    (0,0,0) \pa (0,-6,0) \pb
                    (0,0,0) \pb (4,0,-4) \pc
                    (0,0,0) \pc (0,6,0) \pd
                    (0,0,0) \pd (-4,0,4) \pa
                    };
\end{axis};

\\
};

\node[above] (a2) at ($ (plot1.south west) !0.1! (plot1.south east) $) {(a)};
\node[above] (b2) at ($ (plot2.south west) !0.1! (plot2.south east) $) {(b)};
\node[above] (c2) at ($ (plot3.south west) !0.1! (plot3.south east) $) {(c)};
\node[above] (d2) at ($ (plot4.south west) !0.1! (plot4.south east) $) {(d)};

\end{tikzpicture}

    \caption{An end effector is driven by the parallel combination of two pairs of FREEs with opposing chirality.
    The zonotope (grey areas) is the range of forces that can be produced by applying strictly positive pressure to the individual FREEs.
    It is spanned by the individual force vectors that each FREE produces at maximal pressure (plotted here in the color corresponding to the FREE's appearance in the diagram above).
    By constructing the zonotope for (a) one FREE, (b) two FREEs, (c) three FREEs, (d) four FREEs in parallel, one can observe that all four actuators are needed to gain full control authority over forces along and torques about the $z$-axis.
    In this poorly designed system (with fiber angles and attachments points as shown in the top diagram), the theoretical minimum of 3 actuators is not sufficient to achieve full control authority.}
    \label{fig:zntpConstructed}
\end{figure}

To illustrate the design utility of the force zonotope, consider the parallel combination of FREEs shown in Figure~\ref{fig:zntpConstructed}.
Here, pairs of FREEs with opposite chirality are connected to the two sides of an end effector.
This end effector is constrained to slide and rotate exclusively along and about its $z$-axis.
Since the motion of the end-effector is two-dimensional, we would like to achieve controllable forces within these two dimensions; that is, have control authority over $F^{\vec{z}_e}$ and $M^{\vec{z}_e}$.
When constructing the zonotope one FREE at time (Figure~\ref{fig:zntpConstructed}a-d), one can observe how all four FREEs are needed to achieve this control authority.
In particular, it becomes evident that to achieve full control in $n$ dimensions, at least $n+1$ individual FREEs are required in an antagonistic configuration.
The additional FREE is needed since these soft actuators cannot be driven with negative pressure and can hence not produce bidirectional forces.
One can also observe that if the FREEs are chosen or arranged poorly such that the directions in $\bar{J}_{x}$ do not cover the space of desired forces (such as in Fig.~\ref{fig:zntpConstructed}c), this minimum number of actuators might not be sufficient.

\section{Experimental Evaluation}
\label{sec:experiment}
\begin{figure}
    \centering
    \includegraphics[width=0.75\linewidth]{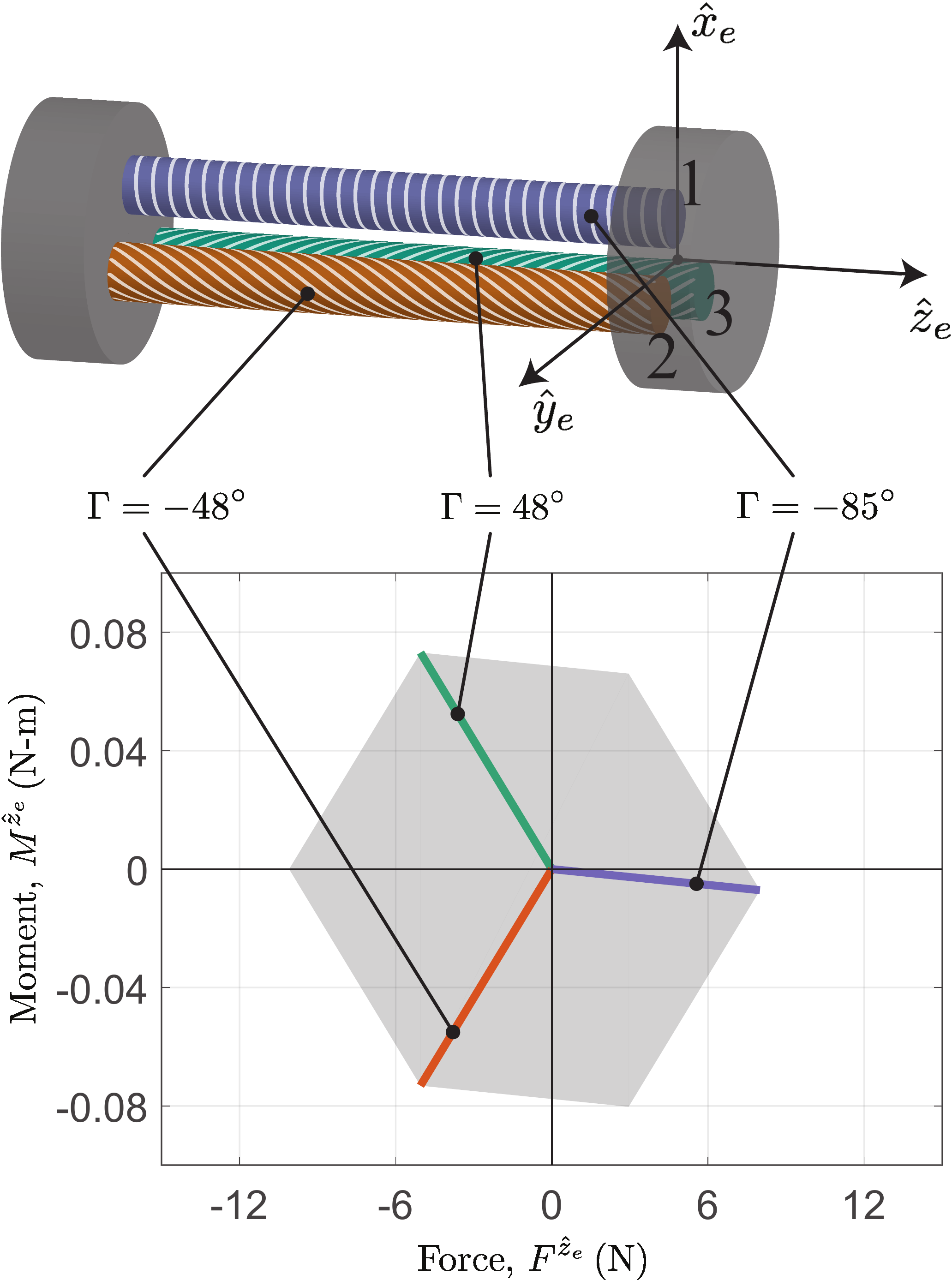}
    \caption{In the experimental evaluation, we employed a parallel combination of three FREEs (top) to yield forces along and moments about the $z$-axis of an end effector.
    The FREEs were carefully selected to yield a well-balanced force zonotope (bottom) to gain full control authority over $F^{\hat{z}_e}$ and $M^{\hat{z}_e}$.
    To this end, we used one extending FREE, and two contracting FREEs which generate antagonistic moments about the end effector $z$-axis.}
    \label{fig:rigDiagram}
\end{figure}

To demonstrate the viability of our modeling methodology, we show experimentally how through the deliberate combination and configuration of parallel FREEs, full control over 2DOF spacial forces can be achieved by using only the minimum combination of three FREEs.
To this end, we carefully chose the fiber angle $\Gamma$ of each of these actuators to achieve a well-balanced force zonotope (Fig.~\ref{fig:rigDiagram}).
We combined a contracting and counterclockwise twisting FREE with a fiber angle of $\Gamma = 48^\circ$, a contracting and clockwise twisting FREE with $\Gamma = -48^\circ$, and an extending FREE with $\Gamma = -85^\circ$.
All three FREEs were designed with a nominal radius of $R$ = \unit[5]{mm} and a length of $L$ = \unit[100]{mm}.

\subsection{Experimental Setup}
\begin{figure}
    \centering
    \includegraphics[width=0.9\linewidth]{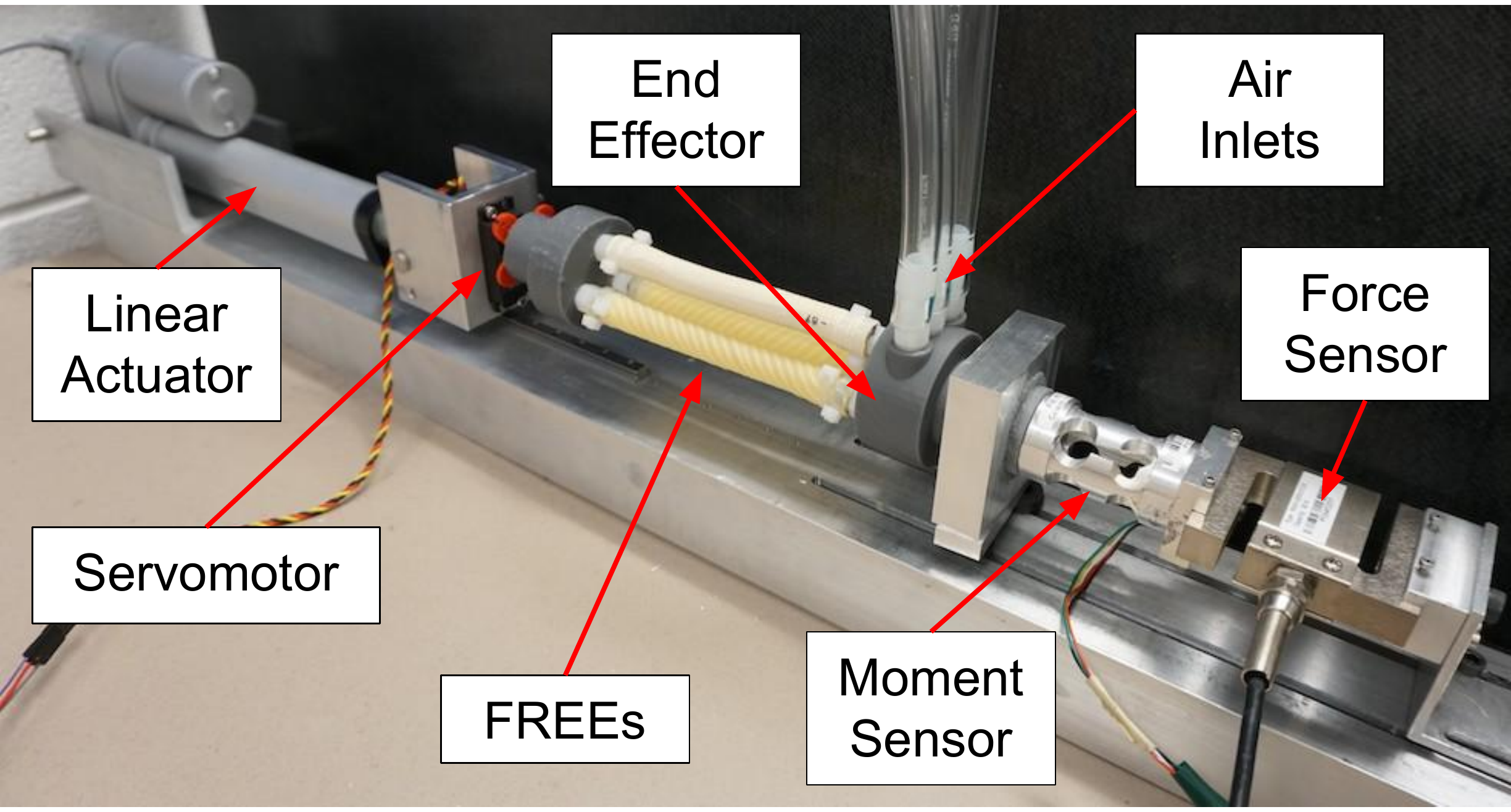}
    \caption{\revcomment{1.3}{This experimental platform is used to generate a targeted displacement (extension and twist) of the end effector and to measure the forces and torques created by a parallel combination of three FREEs. A linear actuator and servomotor impose an extension and a twist, respectively, while the net force and moment generated by the FREEs is measured with a force load cell and moment load cell mounted in series.}}
    \label{fig:rig}
\end{figure}

To measure the forces generated by this actuator combination under a varying state $\vec{x}$ and pressure input $\vec{p}$, we developed a custom built test platform (Fig.~\ref{fig:rig}). 
In the test platform, a linear actuator (ServoCity HDA 6-50) and a rotational servomotor (Hitec HS-645mg) were used to impose a 2-dimensional displacement on the end effector. 
A force load cell (LoadStar  RAS1-25lb) and a moment load cell (LoadStar RST1-6Nm) measured the end-effector forces $F^{\hat{z_e}}$ and moments $M^{\hat{z_e}}$, respectively.
During the experiments, the pressures inside the FREEs were varied using pneumatic pressure regulators (Enfield TR-010-g10-s). 

The FREE attachment points (measured from the end effector origin) were measured to be:
\begin{align}
    \vec{d}_1 &= \bmx 0.013 & 0 & 0 \emx^T  \text{m}\\
    \vec{d}_2 &= \bmx -0.006 & 0.011 & 0 \emx^T  \text{m}\\
    \vec{d}_3 &= \bmx -0.006 & -0.011 & 0 \emx^T \text{m}
%    \vec{d}_i &= \bmx 0 & 0 & 0 \emx^T , && \text{for } i = 1,2,3
\end{align}
All three FREEs were oriented parallel to the end effector $z$-axis:
\begin{align}
    \hat{a}_i &= \bmx 0 & 0 & 1 \emx^T, \hspace{20pt} \text{for } i = 1,2,3
\end{align}
Based on this geometry, the transformation matrices $\bar{\mathcal{D}}_i$ were given by:
\begin{align}
    \bar{\mathcal{D}}_1 &= \bmx 0 & 0 & 1 & 0 & -0.013 & 0 \\ 0 & 0 & 0 & 0 & 0 & 1 \emx^T  \\
    \bar{\mathcal{D}}_2 &= \bmx 0 & 0 & 1 & 0.011 & 0.006 & 0 \\ 0 & 0 & 0 & 0 & 0 & 1 \emx^T  \\
    \bar{\mathcal{D}}_3 &= \bmx 0 & 0 & 1 & -0.011 & 0.006 & 0 \\ 0 & 0 & 0 & 0 & 0 & 1 \emx^T 
%    \bar{\mathcal{D}}_i &= \bmx 0 & 0 & 1 & 0 & 0 & 0 \\ 0 & 0 & 0 & 0 & 0 & 1 \emx^T , && \text{for } i = 1,2,3
\end{align}
These matrices were used in equation \eqref{eq:zeta} to yield the state-dependent fluid Jacobian $\bar{J}_x$ and to compute the resulting force zontopes.
%while using measured values of $\vec{\zeta}^{\,\text{meas}} (\vec{q}, \vec{P})$ and $\vec{\zeta}^{\,\text{meas}} (\vec{q}, 0)$ in \eqref{eq:fiberIso} yields the empirical measurements of the active force.

\subsection{Isolating the Active Force}
To compare our model force predictions (which focus only on the active forces induced by the fibers)
to those measured empirically on a physical system, we had to remove the elastic force components attributed to the elastomer. 
Under the assumption that the elastomer force is merely a function of the displacement $\vec{x}$ and independent of pressure $\vec{p}$ \cite{bruder2017model}, this force component can be approximated by the measured force at a pressure of $\vec{p}=0$. 
That is: 
\begin{align}
    \vec{f}_{\text{elast}} (\vec{x}) = \vec{f}_{\text{\,meas}} (\vec{x}, 0)
\end{align}
With this, the active generalized forces were measured indirectly by subtracting off the force generated at zero pressure:
\begin{align}
    \vec{f} (\vec{x}, \vec{p})  &= \vec{f}_{\text{meas}} (\vec{x}, \vec{p}) - \vec{f}_{\text{meas}} (\vec{x}, 0)     \label{eq:fiberIso}
\end{align}

\subsection{Experimental Protocol}
The force and moment generated by the parallel combination of FREEs about the end effector $z$-axis  was measured in four different geometric configurations: neutral, extended, twisted, and simultaneously extended and twisted (see Table \ref{table:RMSE} for the exact deformation amounts). 
At each of these configurations, the forces were measured at all pressure combinations in the set
\begin{align}
    \mathcal{P} &= \left\{ \bmx \alpha_1 & \alpha_2 & \alpha_3 \emx^T p^{\text{max}} \, : \, \alpha_i = \left\{ 0, \frac{1}{4}, \frac{1}{2}, \frac{3}{4}, 1 \right\} \right\}
\end{align}
with $p^{\text{max}}$ = \unit[103.4]{kPa}. 
\revcomment{3.2}{The experiment was performed twice using two different sets of FREEs to observe how fabrication variability might affect performance. The results from Trial 1 are displayed in Fig.~\ref{fig:results} and the error for both trials is given in Table \ref{table:RMSE}.}

\subsection{Results}

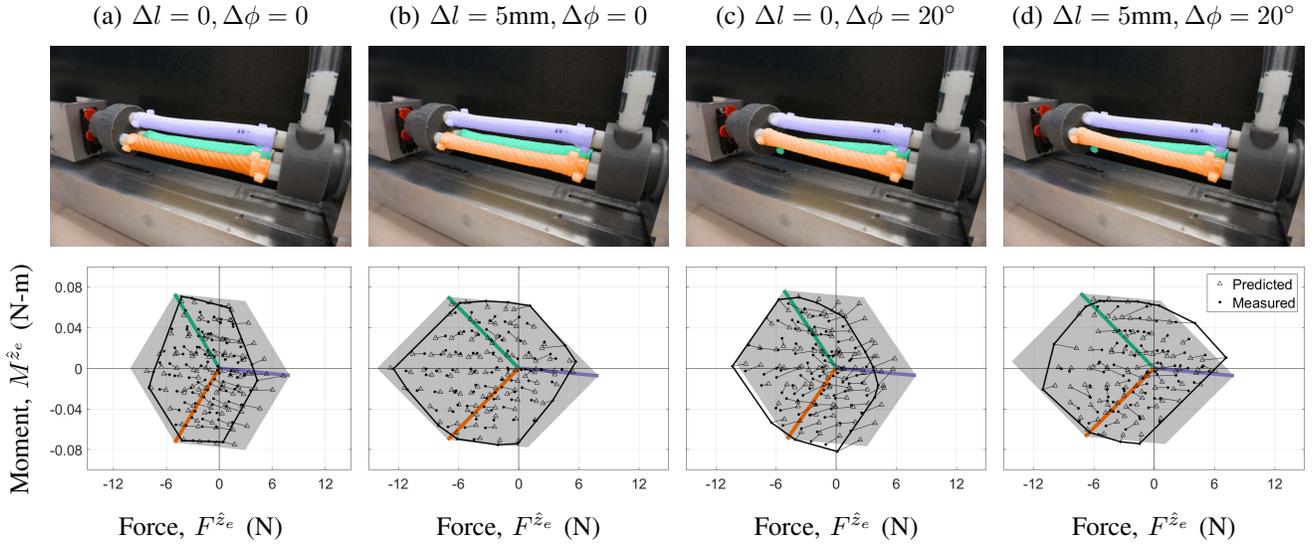
\begin{figure*}[ht]
\centering

\def\picScale{0.08}    % define variable for scaling all pictures evenly
\def\plotScale{0.2}    % define variable for scaling all plots evenly
\def\colWidth{0.22\linewidth}

\begin{tikzpicture} %[every node/.style={draw=black}]
% \draw[help lines] (0,0) grid (4,2);
\matrix [row sep=0cm, column sep=0cm, style={align=center}] (my matrix) at (0,0) %(2,1)
{
& \node (q1) {(a) $\Delta l = 0, \Delta \phi = 0$}; & \node (q2) {(b) $\Delta l = 5\text{mm}, \Delta \phi = 0$}; & \node (q3) {(c) $\Delta l = 0, \Delta \phi = 20^\circ$}; & \node (q4) {(d) $\Delta l = 5\text{mm}, \Delta \phi = 20^\circ$};

\\

&
\node[style={anchor=center}] {\includegraphics[width=\colWidth]{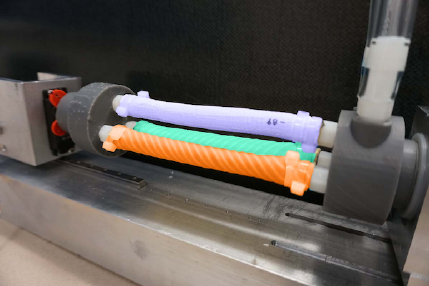}}; %\fill[blue] (0,0) circle (2pt);
&
\node[style={anchor=center}] {\includegraphics[width=\colWidth]{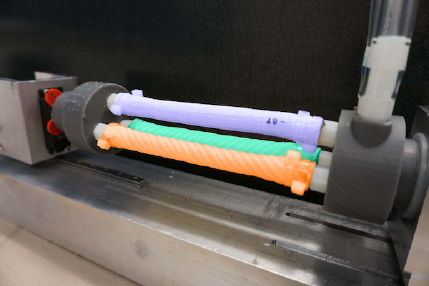}}; %\fill[blue] (0,0) circle (2pt);
&
\node[style={anchor=center}] {\includegraphics[width=\colWidth]{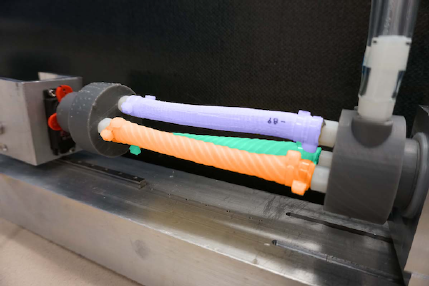}}; %\fill[blue] (0,0) circle (2pt);
&
\node[style={anchor=center}] {\includegraphics[width=\colWidth]{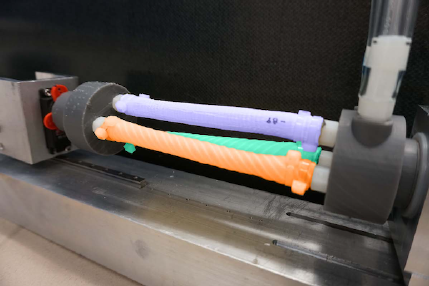}}; %\fill[blue] (0,0) circle (2pt);

\\

\node[rotate=90] (ylabel) {Moment, $M^{\hat{z}_e}$ (N-m)};
&
\node[style={anchor=center}] {\includegraphics[width=\colWidth]{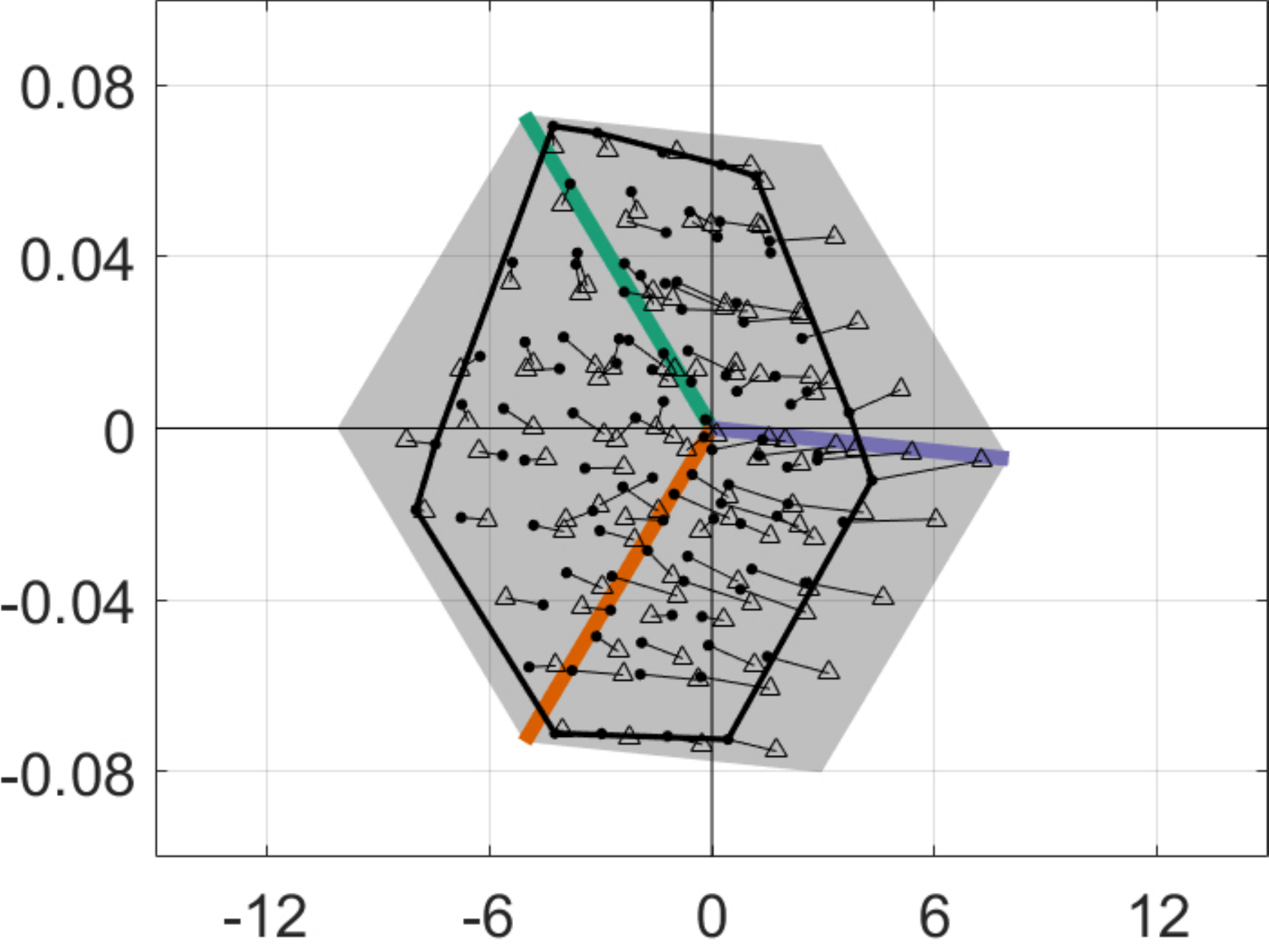}}; %\fill[blue] (0,0) circle (2pt);
&
\node[style={anchor=center}] {\includegraphics[width=\colWidth]{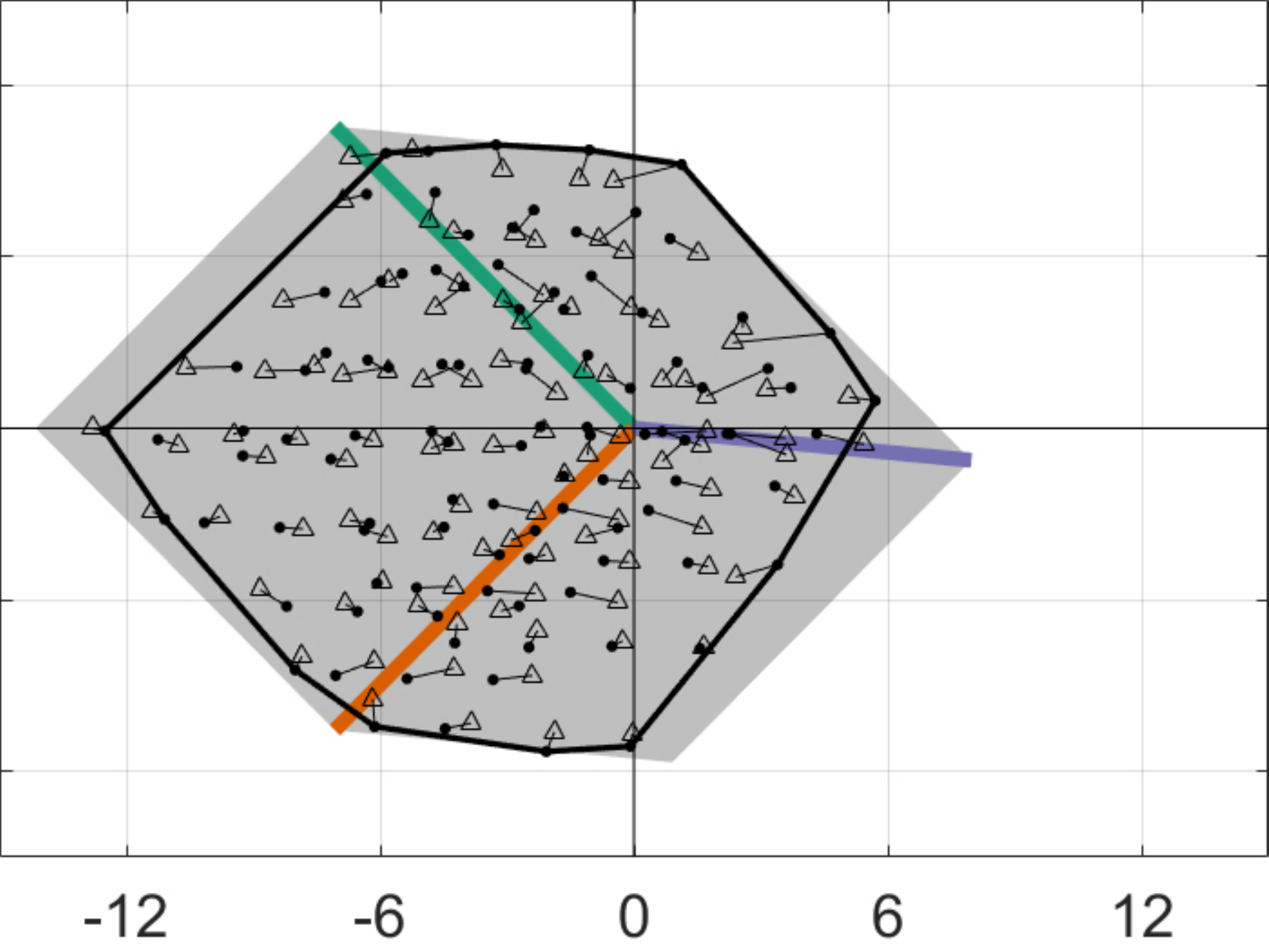}}; %\fill[blue] (0,0) circle (2pt);
&
\node[style={anchor=center}] {\includegraphics[width=\colWidth]{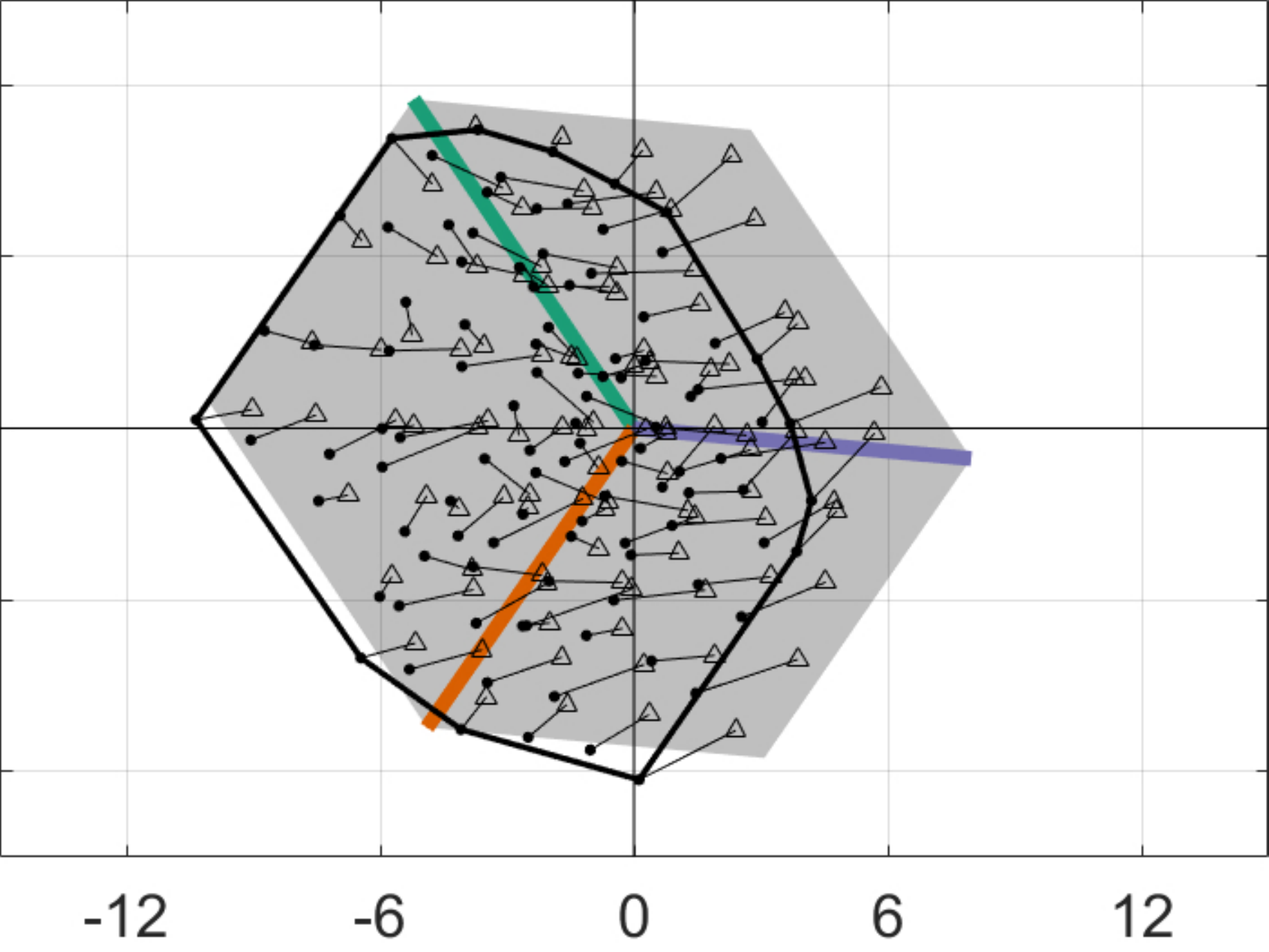}}; %\fill[blue] (0,0) circle (2pt);
&
\node[style={anchor=center}] {\includegraphics[width=\colWidth]{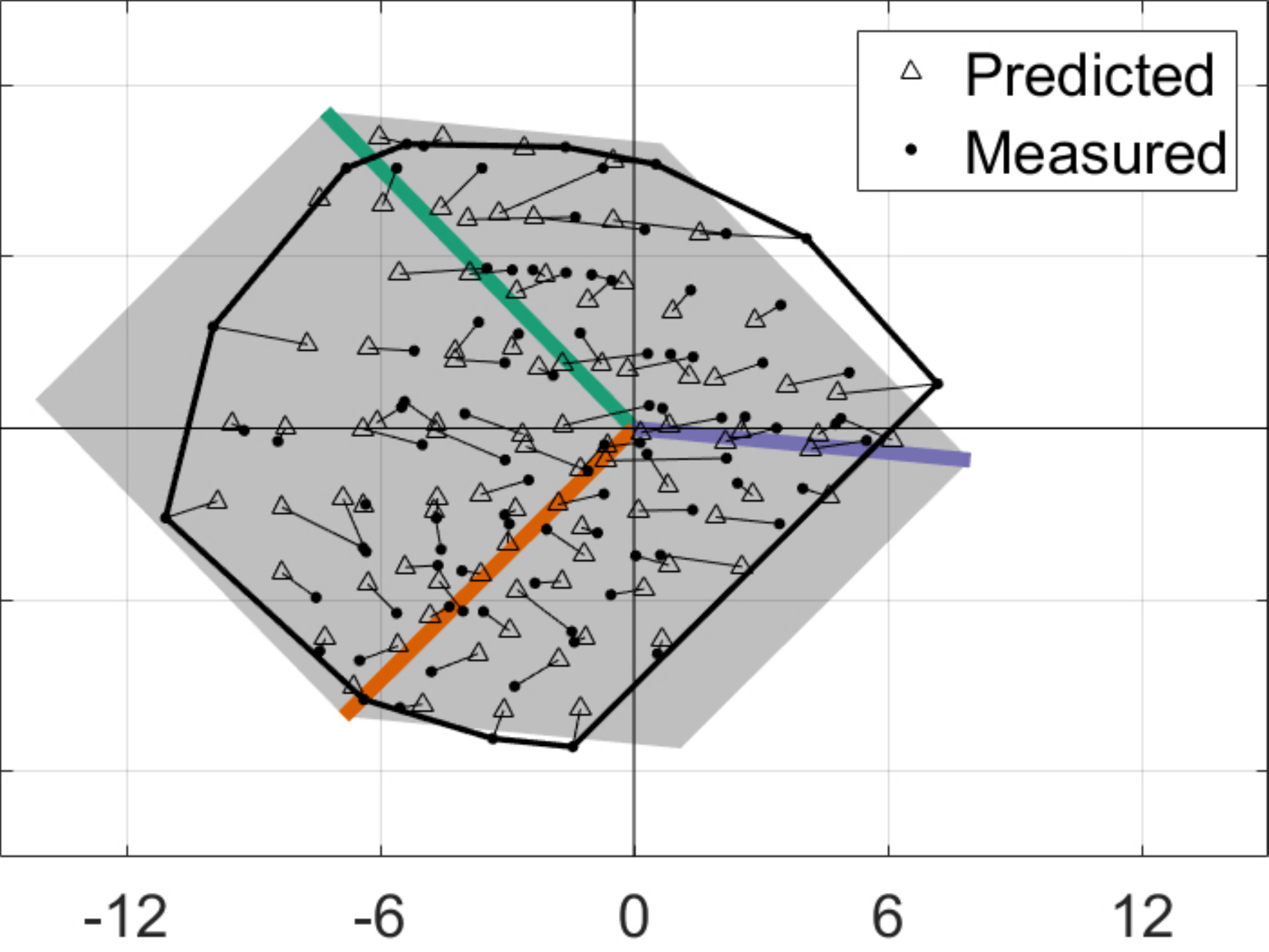}}; %\fill[blue] (0,0) circle (2pt);

\\

& \node (xlabel1) {Force, $F^{\hat{z}_e}$ (N)}; & \node (xlabel2) {Force, $F^{\hat{z}_e}$ (N)}; & \node (xlabel3) {Force, $F^{\hat{z}_e}$ (N)}; & \node (xlabel4) {Force, $F^{\hat{z}_e}$ (N)};

\\
};
\end{tikzpicture}

\caption{For four different deformed configurations (top row), we compare the predicted and the measured forces for the parallel combination of three FREEs (bottom row). 
\revcomment{2.6}{Data points and predictions corresponding to the same input pressures are connected by a thin line, and the convex hull of the measured data points is outlined in black.}
The Trial 1 data is overlaid on top of the theoretical force zonotopes (grey areas) for each of the four configurations.
Identical colors indicate correspondence between a FREE and its resulting force/torque direction.}
\label{fig:results}
\end{figure*}

% & \node (a) {(a)}; & \node (b) {(b)}; & \node (c) {(c)}; & \node (d) {(d)};

For comparison, the measured forces are superimposed over the force zonotope generated by our model in Fig.~\ref{fig:results}a-~\ref{fig:results}d.
To quantify the accuracy of the model, we defined the error at the $j^{th}$ evaluation point as the difference between the modeled and measured forces
\begin{align}
%    \vec{e}_j &= \left( {\vec{\zeta}_{\,\text{mod}}} - {\vec{\zeta}_{\,\text{emp}}} \right)_j
%    e_j &= \left( F/M_{\,\text{mod}} - F/M_{\,\text{emp}} \right)_j
    e^F_j &= \left( F^{\hat{z}_e}_{\text{pred}, j} - F^{\hat{z}_e}_{\text{meas}, j} \right) \\
    e^M_j &= \left( M^{\hat{z}_e}_{\text{pred}, j} - M^{\hat{z}_e}_{\text{meas}, j} \right)
\end{align}
and evaluated the error across all $N = 125$ trials of a given end effector configuration.
% using the following metrics:
% \begin{align}
%     \text{RMSE} &= \sqrt{ \frac{\sum_{j=1}^{N} e_j^2}{N} } \\
%     \text{Max Error} &= \max \{ \left| e_j \right| : j = 1, ... , N \}
% \end{align}
As shown in Table \ref{table:RMSE}, the root-mean-square error (RMSE) is less than \unit[1.5]{N} (\unit[${8 \times 10^{-3}}$]{Nm}), and the maximum error is less than \unit[3]{N}  (\unit[${19 \times 10^{-3}}$]{Nm}) across all trials and configurations.

\begin{table}[H]
\centering
\caption{Root-mean-square error and maximum error}
\begin{tabular}{| c | c || c | c | c | c|}
    \hline
     & \rule{0pt}{2ex} \textbf{Disp.} & \multicolumn{2}{c |}{\textbf{RMSE}} & \multicolumn{2}{c |}{\textbf{Max Error}} \\ 
     \cline{2-6}
     & \rule{0pt}{2ex} (mm, $^\circ$) & F (N) & M (Nm) & F (N) & M (Nm) \\
     \hline
     \multirow{4}{*}{\rotatebox[origin=c]{90}{\textbf{Trial 1}}}
     & 0, 0 & 1.13 & $3.8 \times 10^{-3}$ & 2.96 & $7.8 \times 10^{-3}$ \\
     & 5, 0 & 0.74 & $3.2 \times 10^{-3}$ & 2.31 & $7.4 \times 10^{-3}$ \\
     & 0, 20 & 1.47 & $6.3 \times 10^{-3}$ & 2.52 & $15.6 \times 10^{-3}$\\
     & 5, 20 & 1.18 & $4.6 \times 10^{-3}$ & 2.85 & $12.4 \times 10^{-3}$ \\  
     \hline
     \multirow{4}{*}{\rotatebox[origin=c]{90}{\textbf{Trial 2}}}
     & 0, 0 & 0.93 & $6.0 \times 10^{-3}$ & 1.90 & $13.3 \times 10^{-3}$ \\
     & 5, 0 & 1.00 & $7.7 \times 10^{-3}$ & 2.97 & $19.0 \times 10^{-3}$ \\
     & 0, 20 & 0.77 & $6.9 \times 10^{-3}$ & 2.89 & $15.7 \times 10^{-3}$\\
     & 5, 20 & 0.95 & $5.3 \times 10^{-3}$ & 2.22 & $13.3 \times 10^{-3}$ \\  
     \hline
\end{tabular}
\label{table:RMSE}
\end{table}

\begin{figure}
    \centering
    \includegraphics[width=\linewidth]{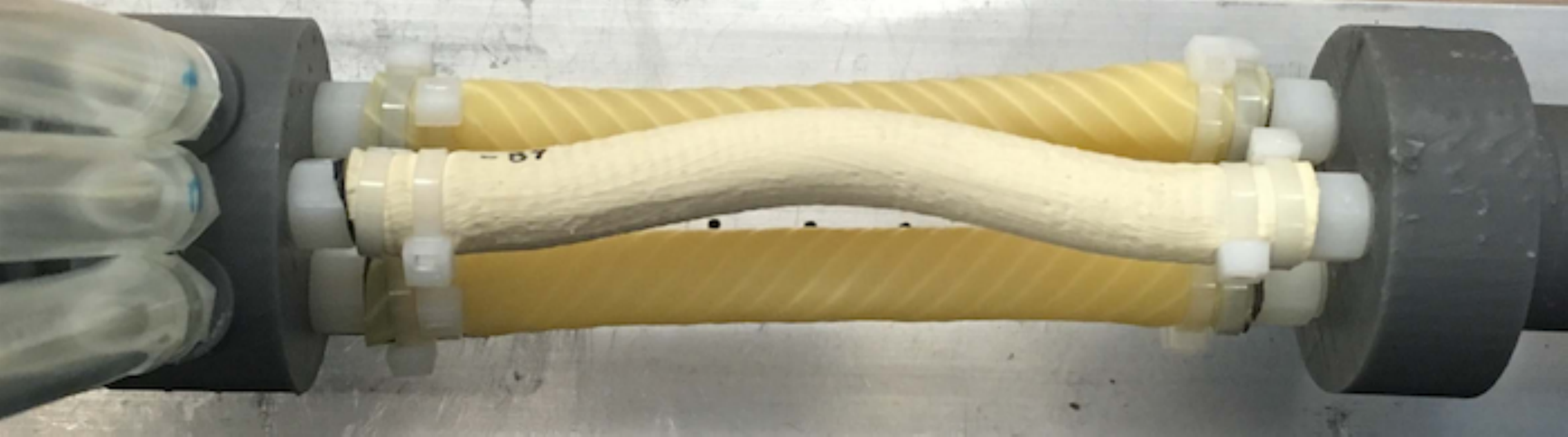}
    \caption{At high fluid pressure the FREE with fiber angle of $-85^\circ$ started to buckle.  This effect was less pronounced when the system was extended along the $z$-axis.}
    \label{fig:buckling}
\end{figure}

%Experimental precision was limited by unmodeled material defects in the FREEs, as well as sensor inaccuracy. While the commercial force and moment sensors used have a quoted accuracy of 0.02\% for the force sensor and 0.2\% for the moment sensor (LoadStar Sensors, 2015), a drifting of up to 0.5 N away from zero was noticed on the force sensor during testing.

It should be noted, that throughout the experiments, the FREE with a fiber angle of $-85^\circ$ exhibited noticeable buckling behavior at pressures above $\approx$ \unit[50]{kPa} (Fig.~\ref{fig:buckling}). 
This behavior was more pronounced during testing in the non-extended configurations (Fig.~\ref{fig:results}a~and~\ref{fig:results}c). 
The buckling might explain the noticeable leftward offset of many of the points in Fig.~\ref{fig:results}a and Fig.~\ref{fig:results}c, since it is reasonable to assume that buckling reduces the efficacy of of the FREE to exert force in the direction normal to the force sensor. 

\begin{figure}
    \centering
    \includegraphics[width=\linewidth]{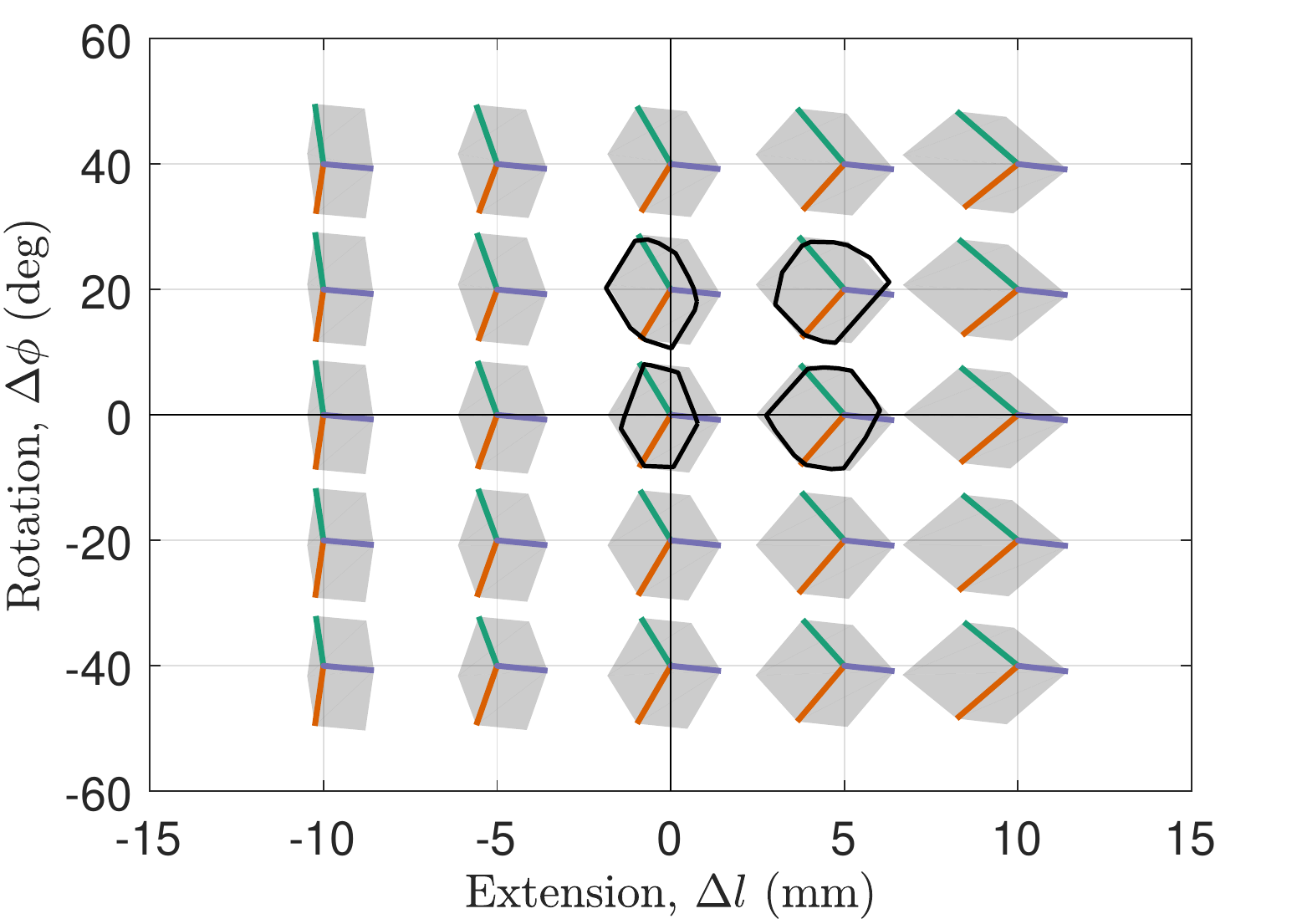}
    \caption{A visualization of how the \emph{force zonotope} of the parallel combination of three FREEs (see Fig.~\ref{fig:rig}) changes as a function of the end effector state $x$. One can observe that the change in the zonotope ultimately limits the work-space of such a system.  In particular the zonotope will collapse for compressions of more than \unit[-10]{mm}.  For \revcomment{2.5}{scale and comparison, the convex hulls of the measured points from Fig.~\ref{fig:results}} are superimposed over their corresponding zonotope at the configurations that were evaluated experimentally.}
    % \marginnote{\#2.5}
    \label{fig:zntp_vs_x}
\end{figure}
\section{Discussion and Conclusions}
\label{sec:conclusion}
In this paper, we present a novel methodology to predict the spatial forces generated by a class of fiber-reinforced fluid-driven actuators (FREEs) that are wound with only one family of fibers.
Our approach is based on the notion of a \emph{fluid Jacobian}, which we propose as the multidimensional and soft equivalent of the cross section $A$ of a traditional pneumatic or hydraulic cylinder.
We use this new modeling approach to predict and visualize the force generation capabilities of parallel combinations of such actuators to yield controllable multidimensional end effector forces.
%The model, derived from energy conservation of a fluid volume constrained by inextensible elements and the superposition of forces acting on a common end effector, predicts the forces generated by a parallel combination of FREEs based on the internal pressures of the actuators and the position of the end effector. 
Our approach is verified experimentally, predicting the forces and moments generated by a parallel combination of three specificially configured FREEs within \unit[3]{N} and \unit[$19 \times 10^{-3}$]{Nm}, respectively.

We envision that the force zonotope presented becomes an instrumental tool in the design of soft robotic systems.
\revcomment{3.5}{This approach could be used to choose a suitable arrangement of FREEs to actuate a compliant manipulator arm, soft orthotic device, or any other application where a custom force profile is desired.}
Here, we used it to systematically arrange the three FREEs in the experimental validation to ensure that the forces and torques created by the individual actuators covered all desired dimensions.
Furthermore, studying the zonotope as a function of state (Fig.~\ref{fig:zntp_vs_x}) allows us to make predictions about the achievable workspace of a soft robotic system.
For example, for the experimental system presented in this work, the zonotope will collapse for compressions of more than \unit[-10]{mm}.
At this point, it will become impossible to create contracting axial forces.

The force zonotope-based model captures the physical behavior of FREEs well enough to be useful in the design of robotic applications.
However, several modeling assumptions limit its accuracy and precision as they might be needed for high-fidelity control. 
Most notably, FREEs are assumed to be cylindrical. 
This assumption introduces inaccuracy when a FREE is bent, buckled, or kinked, \revcomment{2.10}{requiring the development of further models to predict the conditions under which these undesirable behaviors will occur. However, while the model presented does not account for bending or buckling,} 
the fluid Jacobian approach is not inherently limited to cylindrical models and could be extended to more complicated geometrical descriptions of a FREE.
A more fundamental shortcoming is that the model does not explicitly describe the elastomer contributions to force.
That is, our approach must be supplemented with a suitable elastomer model to fully capture the force characteristics of of FREEs.
While this was not the focus of the current work, such a model could be linear based on empirical data \cite{bruder2017model}, derived as a continuum model from first principles \cite{sedal2017constitutive}, or computed via finite element analysis \cite{connolly2015mechanical}.

For soft robots to leverage the advantages in maneuverability, safety, and robustness of non-rigid structures, they require actuators that do not inhibit their compliance. 
A soft actuator such as a FREE meets this criterion because it generates a spacial force without constraining motion to occur in the direction of that force. 
A FREE also has the added benefit of a customizable force direction based on fiber angle; a property that was explicitly exploited in this work.
Parallel combinations of FREEs can be constructed to generate arbitrary spacial forces while still retaining compliance. 
By characterizing the forces generated by parallel combinations of FREEs, this work lays the foundation for future applications of complex soft robotic manipulators.

%                                   % the textheight of the last page by a suitable amount.
%                                   % This command does not take effect until the next page
%                                   % so it should come on the page before the last. Make
%                                   % sure that you do not shorten the textheight too much.

% \input{sections/acknowledgement.tex}

\bibliographystyle{IEEEtran}
\bibliography{references}

\end{document}